\documentclass[conference]{IEEEtran}
\usepackage{amsmath,amssymb,amsfonts}
\newcommand{\minus}{\scalebox{0.75}[1.0]{$-$}}

\usepackage{graphicx,soul}
\usepackage{tabularx}
\usepackage{pgfplots}
\usepackage{makecell}
\usepackage{multicol}
\usepackage{multirow}
\usepackage{subcaption}
\usepackage{algorithm}
\usepackage{pgfplots}
\usepackage{algpseudocode}
\usepackage{esvect}
\usepackage{enumitem}
\usepackage{xcolor}

\setstcolor{red}

\ifCLASSINFOpdf
\else
\fi
\begin{document}
\raggedbottom

\title{Unsupervised Parameter-free Outlier Detection using HDBSCAN* Outlier Profiles}


\makeatletter
\newcommand{\linebreakand}{%
  \end{@IEEEauthorhalign}
  \hfill\mbox{}\par
  \mbox{}\hfill\begin{@IEEEauthorhalign}
}
\makeatother

\author{\IEEEauthorblockN{Kushankur Ghosh}
\IEEEauthorblockA{\textit{Department of Computing Science} \\
\textit{University of Alberta}\\
Edmonton, Canada \\
kushanku@ualberta.ca}
\and
\IEEEauthorblockN{Murilo Coelho Naldi}
\IEEEauthorblockA{\textit{Department of Computing Science} \\
\textit{Federal University of São Carlos}\\
São Paulo, Brazil \\
naldi@ufscar.br}
\and
\IEEEauthorblockN{Jörg Sander}
\IEEEauthorblockA{\textit{Department of Computing Science} \\
\textit{University of Alberta}\\
Edmonton, Canada \\
jsander@ualberta.ca}
\linebreakand
\IEEEauthorblockN{Euijin Choo}
\IEEEauthorblockA{\textit{Department of Computing Science} \\
\textit{University of Alberta}\\
Edmonton, Canada \\
euijin@ualberta.ca}}

\label{tex:title}
\maketitle

\begin{abstract}
In machine learning and data mining, outliers are data points that significantly differ from the dataset and often introduce irrelevant information that can induce bias in its statistics and models.
Therefore, unsupervised methods are crucial to detect outliers if there is limited or no information about them.
\emph{Global-Local Outlier Scores based on Hierarchies} (GLOSH)
is an unsupervised outlier detection method within HDBSCAN*, a state-of-the-art hierarchical clustering method.
GLOSH estimates outlier scores 
for each data point by comparing its density to the highest density of the region they reside in the HDBSCAN* hierarchy. 
GLOSH may be sensitive to HDBSCAN*'s $min_{pts}$ parameter that influences density estimation. 
With limited knowledge about the data, choosing an appropriate $min_{pts}$ value beforehand is challenging as one or some $min_{pts}$ values may better represent the underlying cluster structure than others.
Additionally, in the process of searching for ``potential outliers'', one has to define the number of outliers $n$ a dataset has, which may be impractical and is often unknown.
In this paper, we propose an unsupervised strategy to find the ``best'' $min_{pts}$ value, leveraging the range of GLOSH scores across $min_{pts}$ values to identify the value for which GLOSH scores can best identify outliers from the rest of the dataset. 
Moreover, we propose an unsupervised strategy to estimate a threshold for classifying points into inliers and (potential) outliers without the need to pre-define any value. 
Our experiments show that our strategies can automatically find the $min_{pts}$ value and threshold that yield the best or near best outlier detection results using GLOSH.

\end{abstract}

\begin{IEEEkeywords}
Outlier Detection, Clustering, Parameter Estimation, HDBSCAN*
\end{IEEEkeywords}
\label{tex:abstract}

\IEEEpeerreviewmaketitle



\section{Introduction}
\label{sec:intro}

An outlier is an observation that deviates significantly 
from other observations as to arouse suspicions that it was probably generated by a different mechanism \cite{hawkins1980identification}.
Outliers are common in real-world datasets and often do not contribute to the discovery of knowledge about the data, degrading the 
performance 
of machine learning models.
Sometimes, outliers can exhibit unique behaviors that eventually give insights into areas such as intrusion detection \cite{yuan2021time}.
Hence, detecting outliers is important in machine learning and data mining. 

Outlier detection finds deviating instances using supervised, unsupervised, or one-class classification (OCC) strategies. In a supervised setting, there are enough labeled examples of outliers and inliers 
to train a binary classifier (typically with an imbalance between the number of outliers and the number of inlier examples) \cite{bellinger2012one}. 
In an OCC setting, we have information about inliers and little or no information about outliers \cite{marques2023evaluation}, and a model is learned solely based on information about inliers. 
In an unsupervised setting, there is no prior knowledge about the data, and unsupervised methods identify outliers as instances that deviate from 
a reference set of data.

\noindent \textbf{Motivation.} \emph{Global-Local Outlier Scores based on Hierarchies} (GLOSH) is an unsupervised outlier detection method which is a part of the HDBSCAN* clustering framework \cite{campello2015hierarchical}. It can detect the data points that deviate from their local neighborhood (so-called \emph{local} outliers) and also the data points that
differ more globally from the rest of the data
(so-called \emph{global} outliers).
The GLOSH score of a data point $p$ is computed as the normalized difference between the density 
estimated around $p$
and the highest density estimated in the cluster closest to $p$ in the HDBSCAN* hierarchy.
The hierarchy represents clusters formed at different density levels, and it is constructed w.r.t. a parameter value $min_{pts}$, a smoothing factor
set by a user, that determines the density estimates and, thereby,
the GLOSH score.
The GLOSH score quantifies the outlierness of a data point w.r.t. its closest cluster.
However, different $min_{pts}$ values can yield different HDBSCAN* hierarchies, and one of them may better represent the intrinsic cluster structure of the data than others.
An unsuitable $min_{pts}$ value can result in outliers getting a low GLOSH score.
It has been shown that one may need to use multiple values of $min_{pts}$ to reveal all the clusters in a dataset \cite{cavalcante2021framework}. This means that the closest clusters of a data point could be different depending on the value of $min_{pts}$, and hence a variety of outlier detection results are possible for different $min_{pts}$ values.
In practice, the underlying data distribution is unknown. Therefore, 
choosing a single $min_{pts}$ value that allows the 
detection of most  (preferably all) ``true outliers''  from ``true inliers'' may be challenging.

Moreover, it is essential to pre-define a number \emph{n} of data points with the highest GLOSH scores.
These \emph{n} data points are treated as ``potential outliers'' and are subsequently presented to the domain experts for labeling the ``true outliers'' \cite{zuo2023novel}.
In practice, the number of outliers in a dataset is unknown, which makes it challenging to pick a suitable value for \emph{n} beforehand.

\noindent \textbf{Contribution.} In this paper, we tackle the challenge of choosing a $min_{pts}$ and \emph{n} value that can potentially yield the best results for GLOSH. 
Considering multiple results with increasing $min_{pts}$ values, we show that the GLOSH scores of the data points change at a similar rate at a particular $min_{pts}$ value, resulting in the best overall performance. 
Based on this observation, we designed a strategy based on GLOSH to find the $min_{pts}$ value that yields the best or nearly the best results in a given range of $min_{pts}$ values. At this $min_{pts}$ value, we observed that the distribution of GLOSH scores shows a pattern that we used as a key to develop a strategy that automatically finds a threshold to separate potential outliers from inliers. 
Our strategies can be used in a fully unsupervised way and computed efficiently by running HDBSCAN* following \cite{neto2022core}. In \cite{neto2022core},  it has been shown that 
extracting multiple hierarchies for a range of $min_{pts}$ values is possible with a run-time comparable to running HDBSCAN* twice.
We summarize our major contributions as follows:

\begin{itemize}[nosep]
    \item We introduce the notion of a GLOSH\textendash Profile of a data point \emph{p} as a sequence of GLOSH scores for \emph{p} with respect to a range of $min_{pts}$ values (\ref{sub_sec:method_GLOSH-Profile}). 
    \item We empirically show that scores in a GLOSH\textendash Profile have a pattern that allows the identification of the $min_{pts}$ value that best distinguishes inliers and outliers (\ref{sub_sec:glosh_performance} and \ref{sub_sec:stable_behavior_profiles}).
  \item We introduce automatic parameter selection using GLOSH, called Auto-GLOSH, a
  strategy leveraging GLOSH\textendash Profiles to find the $min_{pts}$ value that leads to the best outlier detection using GLOSH (\ref{sub_sec:selective_profile_method}).
  \item We introduce Potential Outlier Labelling AppRoach (POLAR), a
  strategy 
  leveraging
  the distribution of 
  GLOSH scores 
  at
  the $min_{pts}$ value estimated by Auto-GLOSH to find 
   the threshold for labeling potential outliers (\ref{sub_sec:inlier_potentialOutlier_separation}).
\end{itemize}




\label{tex:intro}

\section{Related Work}
\label{sec:related_work}

Unsupervised outlier detection is based on the premise that potential outliers lie in low-density regions of the reference set. In \cite{knorr1997unified}, outliers are data points beyond a pre-defined distance from a large proportion of data. 
However, finding a distance threshold for explicitly identifying outliers is challenging. 
Therefore, algorithms that compute an outlier score for each data point were developed that do not rely on a fixed distance for labeling outliers. 
 These methods are often categorized into \emph{local} and \emph{global} in the literature \cite{marques2023evaluation}. A classic global method is \emph{k-Nearest Neighbor} ($k\mathrm{NN}$) \cite{ramaswamy2000efficient}, where the outlier score of a data point is
  the distance to its $k^{th}$ nearest neighbor $(k\mathrm{NN\minus dist})$. 
 In $k\mathrm{NN}$ the potential outliers are expected to have large $k\mathrm{NN\minus dist}$ values and 
 can be considered ``global'' as their deviation is measured w.r.t. the overall dataset. On the other hand, the Local Outlier Factor (LOF) \cite{breunig2000lof} is another classic method that finds outliers 
 by assessing the density deviation of
 each data point 
 relative to the deviation of its nearest neighbors (local neighborhood) used as the reference set. These outliers are considered ``local'' as they are outliers w.r.t. their nearest neighbors. Thus, choosing an appropriate method depends on the type of outliers present in the data, which is challenging to know beforehand. Differently, GLOSH proposes a way to dynamically select the reference sets based on the HDBSCAN* hierarchy, offering a more adaptive approach to detect both types of outliers \cite{campello2015hierarchical}.

    However, GLOSH requires the $min_{pts}$ parameter indirectly since it is required by HDBSCAN* to construct the cluster hierarchy, which is known to be  sensitive; even small changes in the $min_{pts}$ value can lead to very different data partitions \cite{ankerst1999optics}. Swersky \cite{swersky2018study} systematically showed that the performance of GLOSH can significantly vary for different $min_{pts}$ values. 
    While automatic methods exist for extracting flat clustering results from hierarchies \cite{ankerst1999optics,campello2013framework}, there is no prior research for the automatic selection of the $min_{pts}$ value for hierarchical clustering methods.
    Finding the best $min_{pts}$ value for GLOSH in combination with HDBSCAN* is challenging in unsupervised scenarios, as there is no prior knowledge about the underlying distribution of the data.
    Recent works such as \cite{zhao2021automatic} uses meta learning to select outlier detection models for a new dataset based on its past performance on similar datasets.
    Different from 
    \cite{zhao2021automatic}, we use a single outlier detection model, GLOSH, and do not select it from a set of models.
    Additionally, our method does not require prior information about 
    the dataset or the past performance of GLOSH on similar datasets. 
    Such prior information, while advantageous, may not always be available.
\label{tex:related_works}

\section{Background}
\label{sec:background}

\subsection{HDBSCAN*}
\label{sub_sec:hdbscan}

HDBSCAN* or \emph{Hierarchical DBSCAN*} \cite{campello2015hierarchical} is a hierarchical way of clustering data
which is an improvement of one of the most popular density-based clustering algorithms, DBSCAN \cite{ester1996density}.
In DBSCAN*, density-based clusters are constructed using two parameters: (I) the radius $\epsilon$, and (II) the minimum number of points 
$min_{pts}$. 
A point $x_i$ in a dataset $D$ 
is a \emph{core point} if it has $min_{pts}$ data points in its $\epsilon$\textendash \emph{neighborhood} $N_\epsilon(x_i)$ (i.e. in the set of points that are located less than or equal to distance $\epsilon$ away from $x_i$); otherwise, a point is labeled as noise. Two \emph{core points} are \emph{density connected} w.r.t. $\epsilon$ and $\min_{pts}$ if they directly or transitively belong to each other's $N_\epsilon$. DBSCAN* defines a cluster as a maximal set of points where each pair of points in the set are \emph{density connected}.


HDBSCAN* improves DBSCAN* and does not require the user to pre-define an $\epsilon$ radius. For a $min_{pts}$ value, 
HDBSCAN* computes (I) the \emph{core distance} $\epsilon_{c}(x_i)$ for each $x_i \in D$, which is the minimum $\epsilon$ distance required for $x_i$ to be a \emph{core point}, and (II) the \emph{mutual reachability distance} $d_{mrd}(x_i,x_j)$, which is the minimum distance required
for $x_i$ and $x_j$ to be in each other's $\epsilon$-neighborhood
while both are also \emph{core points}.
The $d_{mrd}(x_i,x_j)$ 
is defined as:
\raggedbottom
\begin{equation}\label{d_mrd}    
d_{mrd} (x_i, x_j) = max\{\epsilon_{c}(x_i), \epsilon_{c}(x_j), d(x_i,x_j)\}
\end{equation}
\raggedbottom
\noindent where $d(x_i,x_j)$ 
represents the underlying distance between the two points in the dataset. The mutual reachability distance $d_{mrd}$ is dependent on the value of $min_{pts}$. The HDBSCAN* hierarchy w.r.t. a single $min_{pts}$ value is obtained by
computing the \emph{minimum spanning tree} $MST_{min_{pts}}$ of a complete, edge-weighted, virtual graph $G_{min_{pts}}$ \cite{neto2022core}, where the vertices represent the data points in the dataset, and the edge weights are the values of the \emph{mutual reachability distance} $d_{mrd}$ between them. 
The edges from the MST are then removed in decreasing order of edge weights. At every removal, we obtain a new set of connected core objects (which are the clusters) and the remaining noise (points that are not connected). 
This creates a hierarchy of all possible DBSCAN* solutions obtained at $\epsilon \in [0, \infty]$ for a given $min_{pts}$ value. 

\subsection{GLOSH}
\label{sub_sec:glosh}

The HDBSCAN* framework includes the \emph{Global-Local Outlier Scores based on Hierarchies} (GLOSH) algorithm to detect outliers using hierarchical density estimates. GLOSH computes a score for each data point in a dataset w.r.t. a reference set of points dynamically selected from the cluster structure within the HDBSCAN* hierarchy.

The GLOSH score for a point $x_i \in D$ is based on $C_{x_i}$, which is the cluster where $x_i$ is assigned to for the minimum $\epsilon$ value in the hierarchy that makes  $x_i$ a core point.
The density of $x_i$, $\lambda(x_i)$ is compared to that of the densest point in $C_{x_i}$, $\lambda_{max}(C_{x_i})$, which is the point that is assigned to the $C_{x_i}$ when it is first formed in the hierarchy (making it the longest surviving point in $C_{x_i}$). The GLOSH score $\Gamma_{min_{pts}}(x_i)$ of $x_i$ for a particular $min_{pts}$ value is defined as:
\raggedbottom
\begin{equation}
\label{eq:glosh}
    \Gamma_{min_{pts}}(x_i) = \frac{\lambda_{max}(C_{x_i}) - \lambda(x_i)}{\lambda_{max}(C_{x_i})}
\end{equation}
\raggedbottom
\noindent where $\lambda$ is defined as $\frac{1}{\epsilon_{c}(x_i)}$ for any point $x_i$.
The
$\lambda_{max}(C_{x_i})$
is used as the referential density to compare the density of any point $x_i$ that has $C_{x_i}$ as its original cluster in the hierarchy. 
The GLOSH score falls in the range $[0,1)$. If $x_i$ resides in a dense region of a cluster, then $\lambda_{max}(C_{x_i}) - \lambda(x_i)$ tends to be $0$ whereas, for points that reside in low-density areas,  $\lambda_{max}(C_{x_i}) - \lambda(x_i)$ tends to increase, resulting in higher GLOSH scores. 
A pre-defined threshold is applied to the GLOSH scores to label outliers.

Thus, GLOSH depends on two parameters that are often unknown in practice: the $min_{pts}$ for computing the point densities and the threshold for labeling outliers. In section \ref{sec:method}, we propose a way to find the ``best'' $min_{pts}$ value, and in section  \ref{sec:threshold_estimation}, a strategy to determine the threshold.


\label{tex:background}

\section{Using GLOSH\textendash Profiles for $min_{pts}$ Selection}
\label{sec:method}


This section discusses selecting the best $min_{pts}$ value for GLOSH in four parts: 
(1) introducing GLOSH\textendash Profiles as a tool to find the best $min_{pts}$ value (\ref{sub_sec:method_GLOSH-Profile}), (2) linking the uniform rate of change in profiles to the best $min_{pts}$ value (\ref{sub_sec:glosh_performance} and \ref{sub_sec:stable_behavior_profiles}), (3) proposing Auto-GLOSH to automate the selection from the profiles (\ref{sub_sec:selective_profile_method}), and (4) outlining the complexity of Auto-GLOSH (\ref{sub_sec:complexity_analysis}).
\raggedbottom

\subsection{GLOSH\textendash Profile}
\label{sub_sec:method_GLOSH-Profile}


This sub-section introduces the notion of a GLOSH\textendash Profile that captures the behavior of the GLOSH scores of a data point 
across different HDBSCAN* hierarchies over a range of $min_{pts}$ values.
The impact of changing the $min_{pts}$ values on cluster formation has been showed in prior studies \cite{neto2019efficient,neto2022core}.
An increase in $min_{pts}$ values requires each point to have more points in its $\epsilon$\textendash \emph{neighborhood}
    to become a \emph{core point}, resulting in larger core distance $\epsilon_c(.)$ values. 
    In turn, larger core distances tend to form ``smoother'' (and possibly larger) clusters in the hierarchy. Conversely, decreasing $min_{pts}$ values allows points to become \emph{core points} with smaller $\epsilon_c(.)$ values, potentially building more detailed and smaller clusters within the base of the hierarchy. 
    Therefore, considering different $min_{pts}$ values, GLOSH may compare a point's density with that of different clusters (i.e., neighborhoods of different densities), which influences the score for that point.
    Consequently, different $min_{pts}$ values may yield different results, and it is challenging to know which is the best.
    To address the problem of finding a single best value for $min_{pts}$ for GLOSH we formally define a GLOSH\textendash Profile as follows:

\noindent\textbf{Definition 4.1 (GLOSH\textendash Profile).}\label{def:glosh_prof} 
A \emph{GLOSH\textendash Profile} $P\Gamma_{m_{max}}(x_i)$
is an array of GLOSH scores $\Gamma_{min_{pts}}(x_i)$ for a given point $x_i$ for a range $[2, m_{max}]$ of $min_{pts}$ values between $2$ and a maximum value $m_{max}$:
\begin{align}
    P\Gamma_{m_{max}}(x_i) &= \begin{bmatrix}
                       \Gamma_{2}(x_i) \\
                       \Gamma_{3}(x_i) \\
                       \vdots \\
                       \Gamma_{m_{max}}(x_i)
                     \end{bmatrix}
    \end{align}

The $m_{max}$ value is in practice typically chosen to be smaller than $100$ \cite{marques2015internal}. In this paper, when $m_{max}$ is clear from the context or irrelevant for the discussion, we abbreviate $P\Gamma_{m_{max}}(.)$ as $P\Gamma(.)$. We use GLOSH\textendash Profiles as a
guide to select the ``best'' $min_{pts}$ value.

\subsection{GLOSH Performance across $min_{pts}$ values}
\label{sub_sec:glosh_performance}

We analyze the performance of GLOSH across various $min_{pts}$ values to find the specific ranges where GLOSH achieves its best performance. Additionally, we compare these results with the GLOSH\textendash Profiles 
as we aim to find the $min_{pts}$ values that lead to the best GLOSH performance.


\begin{figure}[t]
\centering
\begin{subfigure}{2.25cm}
\centering\includegraphics[width=2.25cm,height=1.6cm]{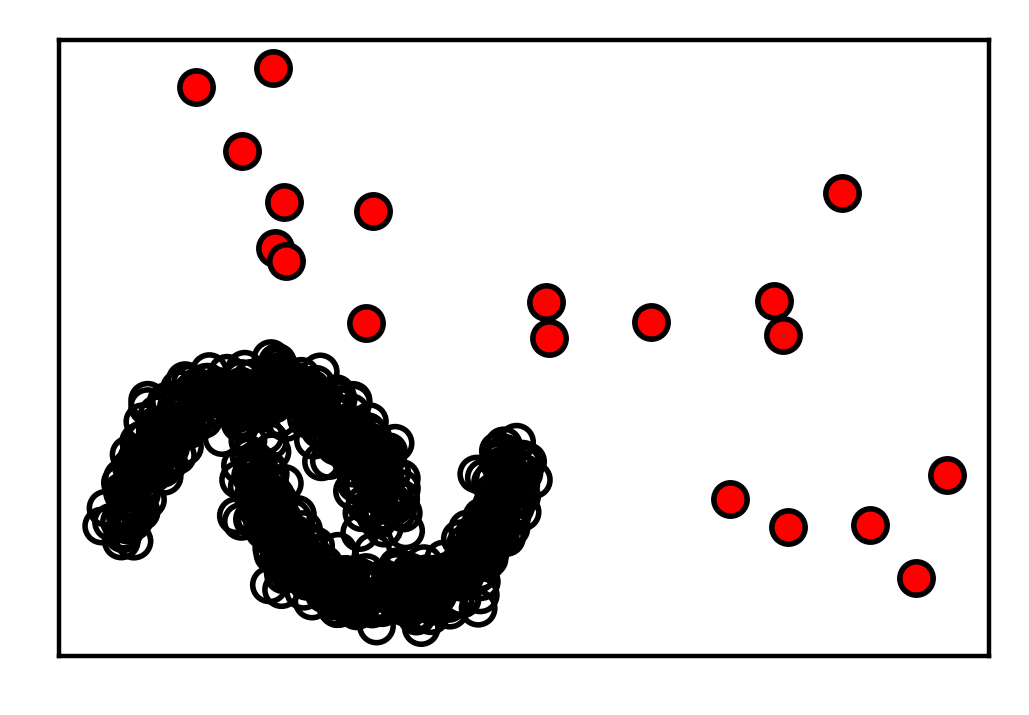}
\caption{Global}
\label{fig:zenodo_banana_global}
\end{subfigure}
\begin{subfigure}{2.25cm}
\centering\includegraphics[width=2.25cm,height=1.6cm]{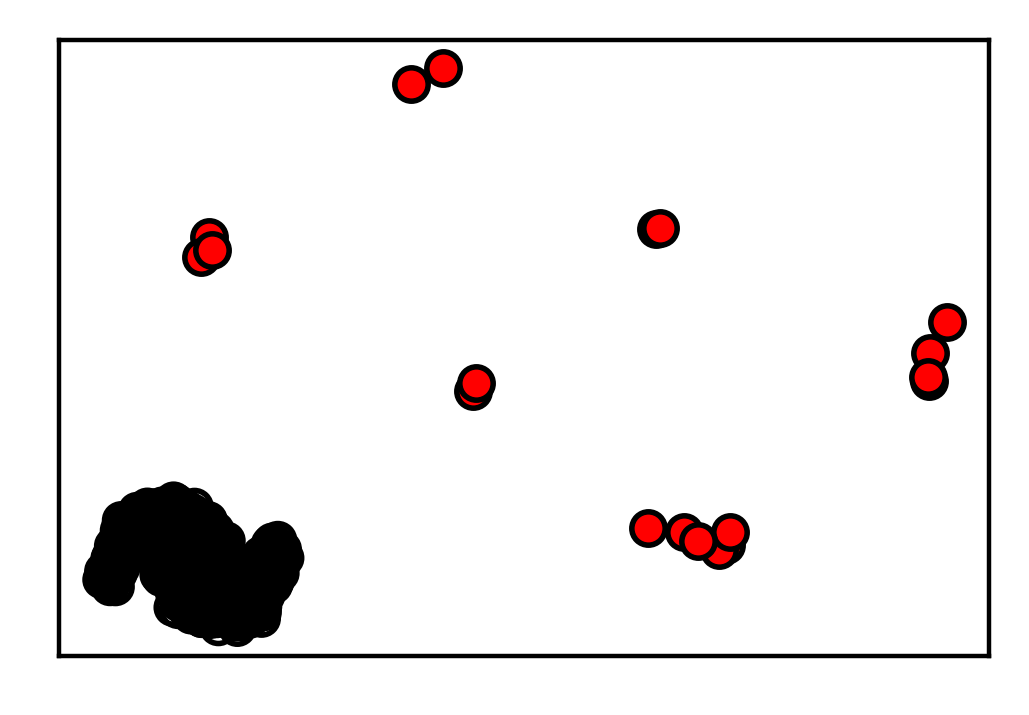}
\caption{Clumps}
\label{fig:zenodo_banana_clump}
\end{subfigure}
\begin{subfigure}{2.25cm}
\centering\includegraphics[width=2.25cm,height=1.6cm]{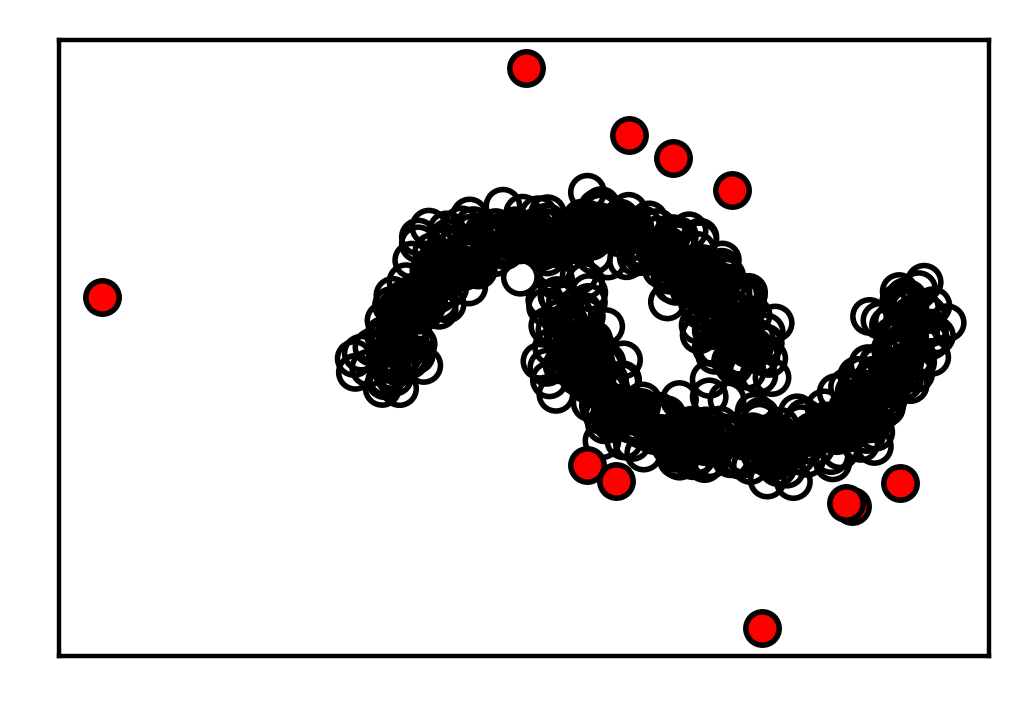}
\caption{Local}
\label{fig:zenodo_banana_local}
\end{subfigure}


\caption{Banana Dataset with different kinds of outliers.}

\label{fig:zenodo_datasets_2d}
\end{figure}
\raggedbottom

The performance of GLOSH is assessed using the Precision@n (P@n) metric; it measures the fraction of ``true outliers'' among the points with the $n$-th highest GLOSH scores, where $n$ is taken as the total number of labeled outliers in a test dataset. 
In this section, we use two-dimensional datasets to allow the visualization of the data distribution, 
 facilitating an intuitive understanding of the results.
We use the dataset with banana-shaped clusters from \cite{koncar_2018_1171077}, commonly used for outlier detection studies in the literature \cite{marques2023evaluation}.
To investigate outliers with different spatial characteristics, 
we followed \cite{han2022adbench} to generate different kinds of synthetic outliers such as local, global, and small groups of points (i.e., “outlier clumps” \cite{marques2020internal}) with similar behavior among themselves but dissimilar from the rest of the data. 
To generate local outliers, we learn a Gaussian Mixture Model (GMM) on the inlier samples. 
The covariance matrix $\Sigma$ of the inlier samples is estimated using the GMM model and is
scaled to $\hat{\Sigma} = \alpha\Sigma$ (where $\alpha = 5$ as in \cite{han2022adbench}). 
We use the GMM model to generate the local outlier samples using $\hat{\Sigma}$. 
To generate outlier clumps, 
we use the GMM to
estimate the mean feature vector $\mu$ of the inlier class samples. Then, we scale the mean feature vector by a factor of $5$ ($\hat{\mu} = \alpha\mu$ where $\alpha = 5$ as in \cite{han2022adbench}) to generate the clumps using the GMM model. 
Hereafter, we will refer to these outliers as `Clumps'. We generate global outliers from a uniform distribution with boundaries defined by $(\alpha \cdot max(D^m), \alpha \cdot min(D^m))$ where $D^m$ represents the $m$-th feature of the inlier class data. Then, we apply the \emph{tomek links} technique \cite{krawczyk2019instance} to filter out generated points that fall "too close" to the existing inlier points, which would not characterize them as global outliers.
 An inlier and outlier data point forms a \emph{tomek link} when they are the closest neighbors to each other.
\emph{Tomek link} indicates that the generated outlier falls inside 
or is located close to the border of a cluster and should be removed. 
Fig. \ref{fig:zenodo_datasets_2d} shows the datasets, including the generated outliers.

\begin{figure}[t]
\centering
\begin{subfigure}{2.3cm}
\centering\includegraphics[width=2.3cm]{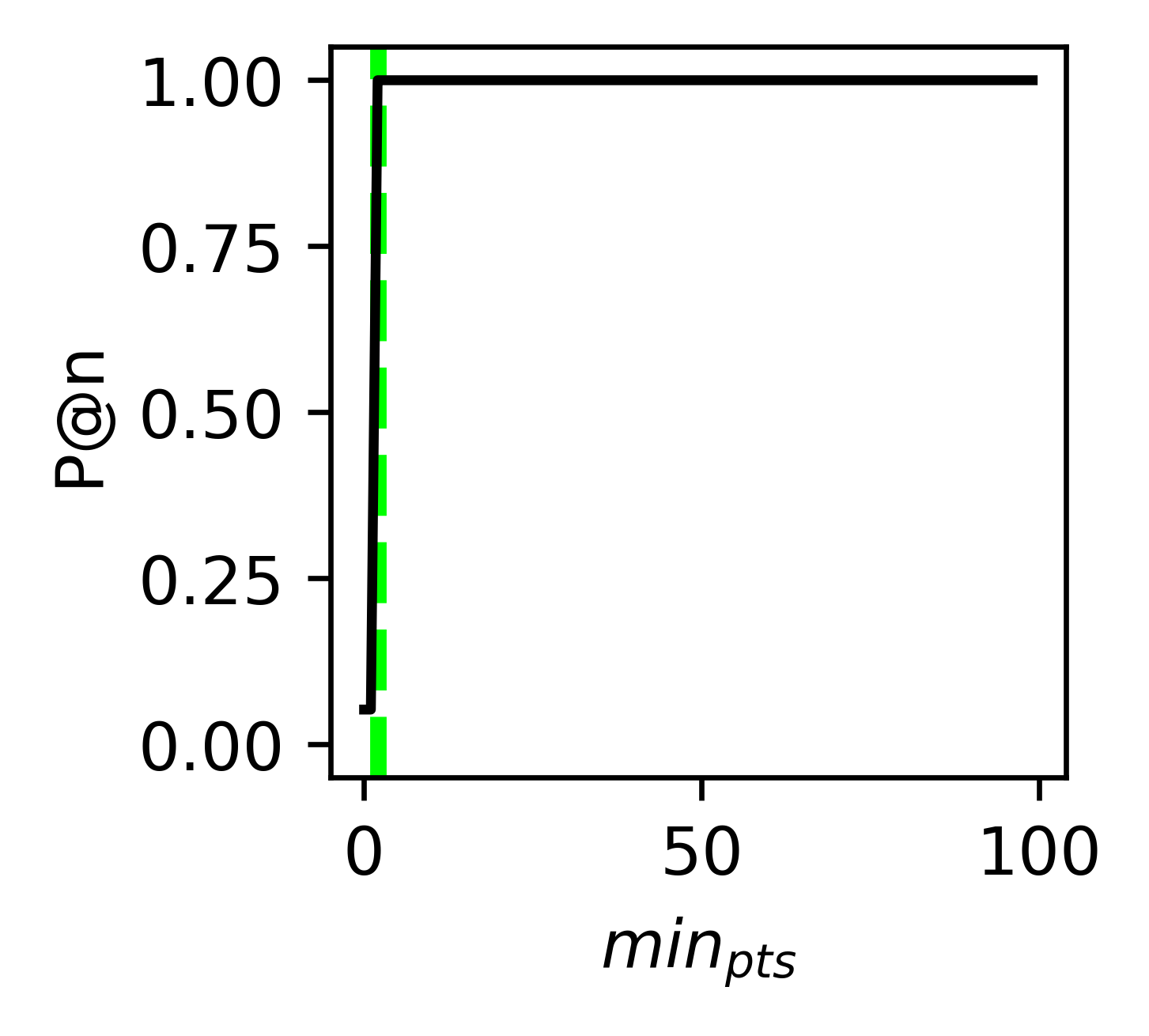}
\caption{Global-P@n}
\label{fig:zenodo_banana_global_prec}
\end{subfigure}
\begin{subfigure}{2.3cm}
\centering\includegraphics[width=2.3cm]{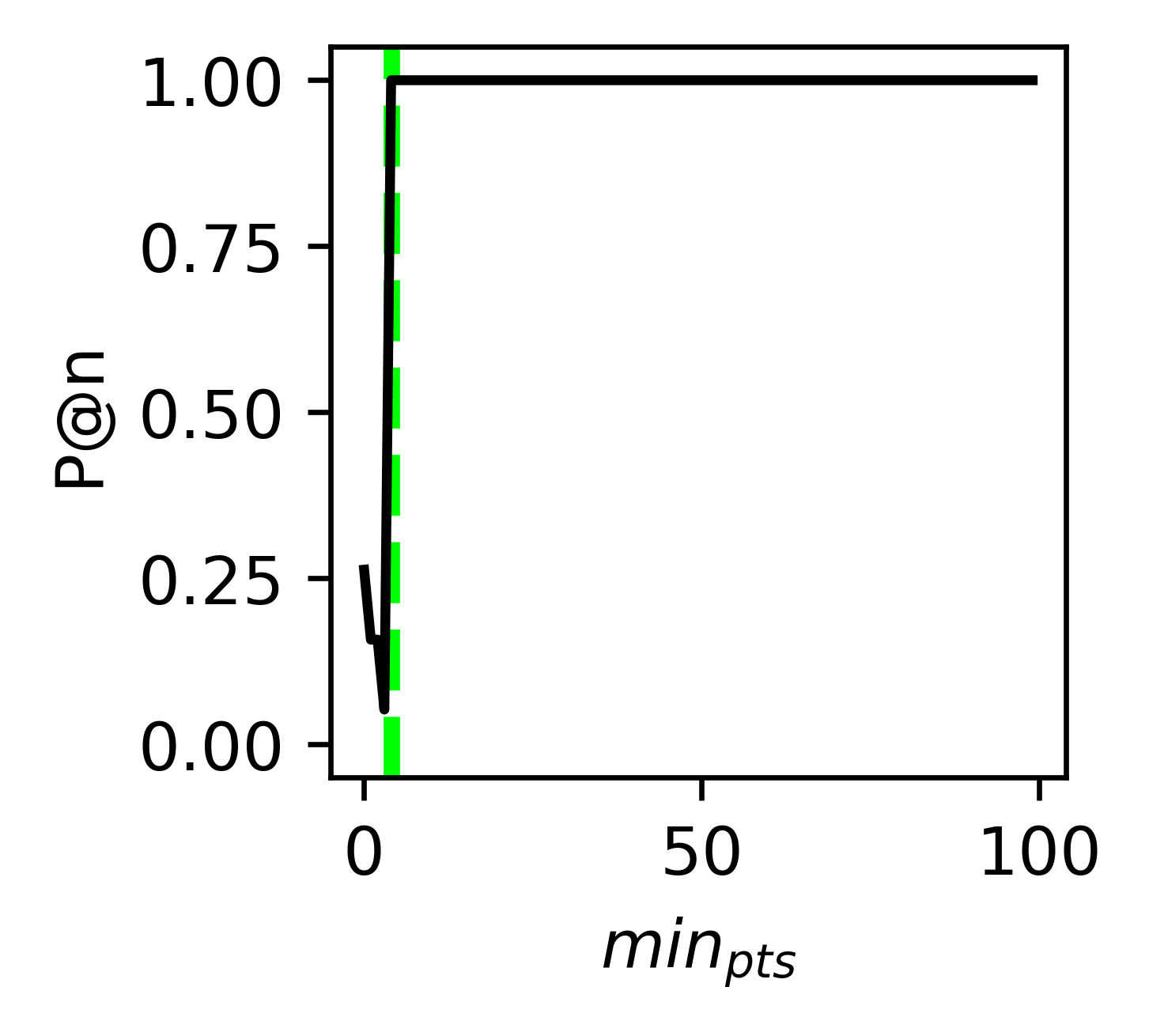}
\caption{Clumps-P@n}
\label{fig:zenodo_banana_clump_prec}
\end{subfigure}
\begin{subfigure}{2.3cm}
\centering\includegraphics[width=2.3cm]{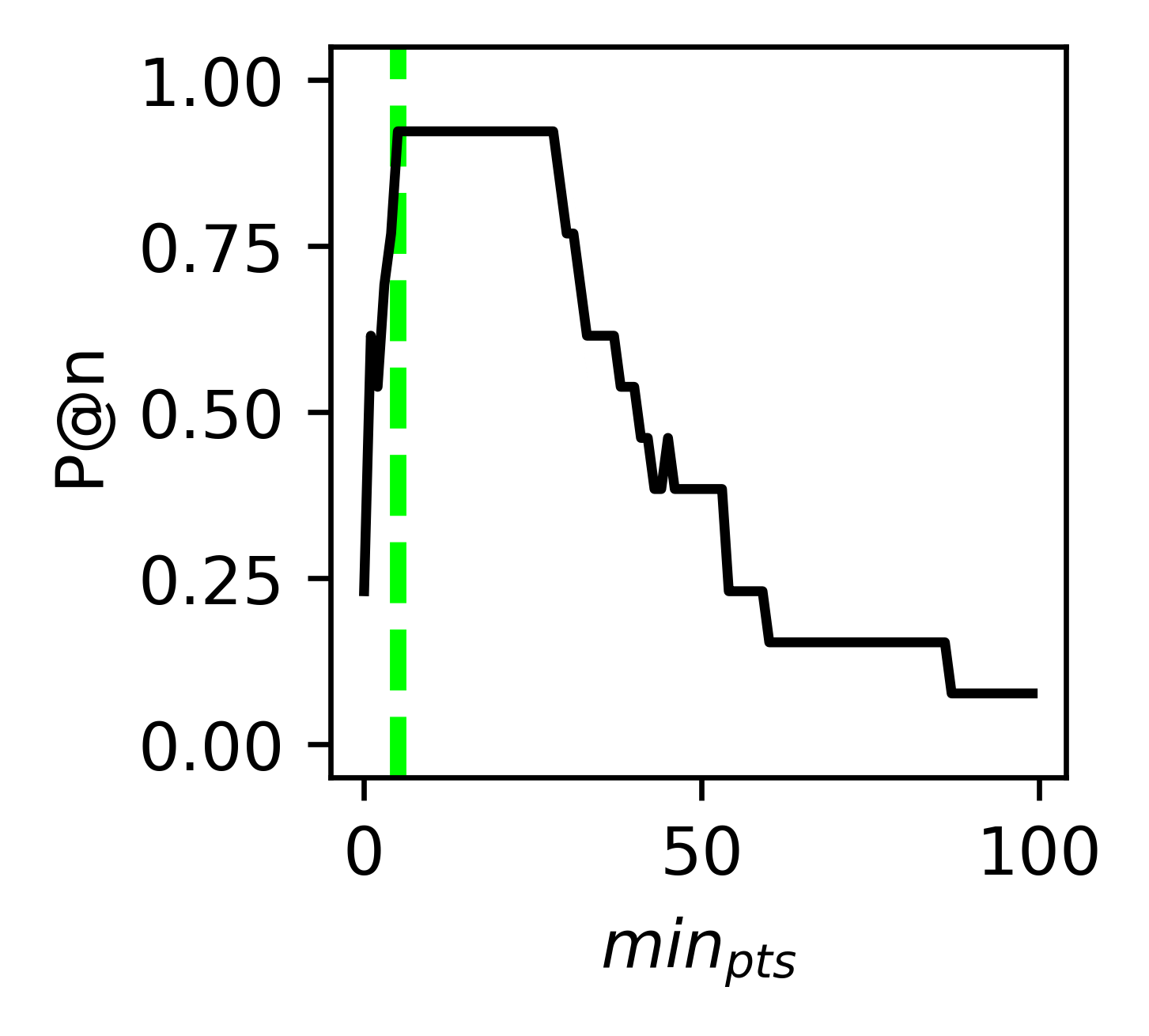}
\caption{Local-P@n}
\label{fig:zenodo_banana_local_prec}
\end{subfigure}

\begin{subfigure}{2.3cm}
\centering\includegraphics[width=2.3cm]{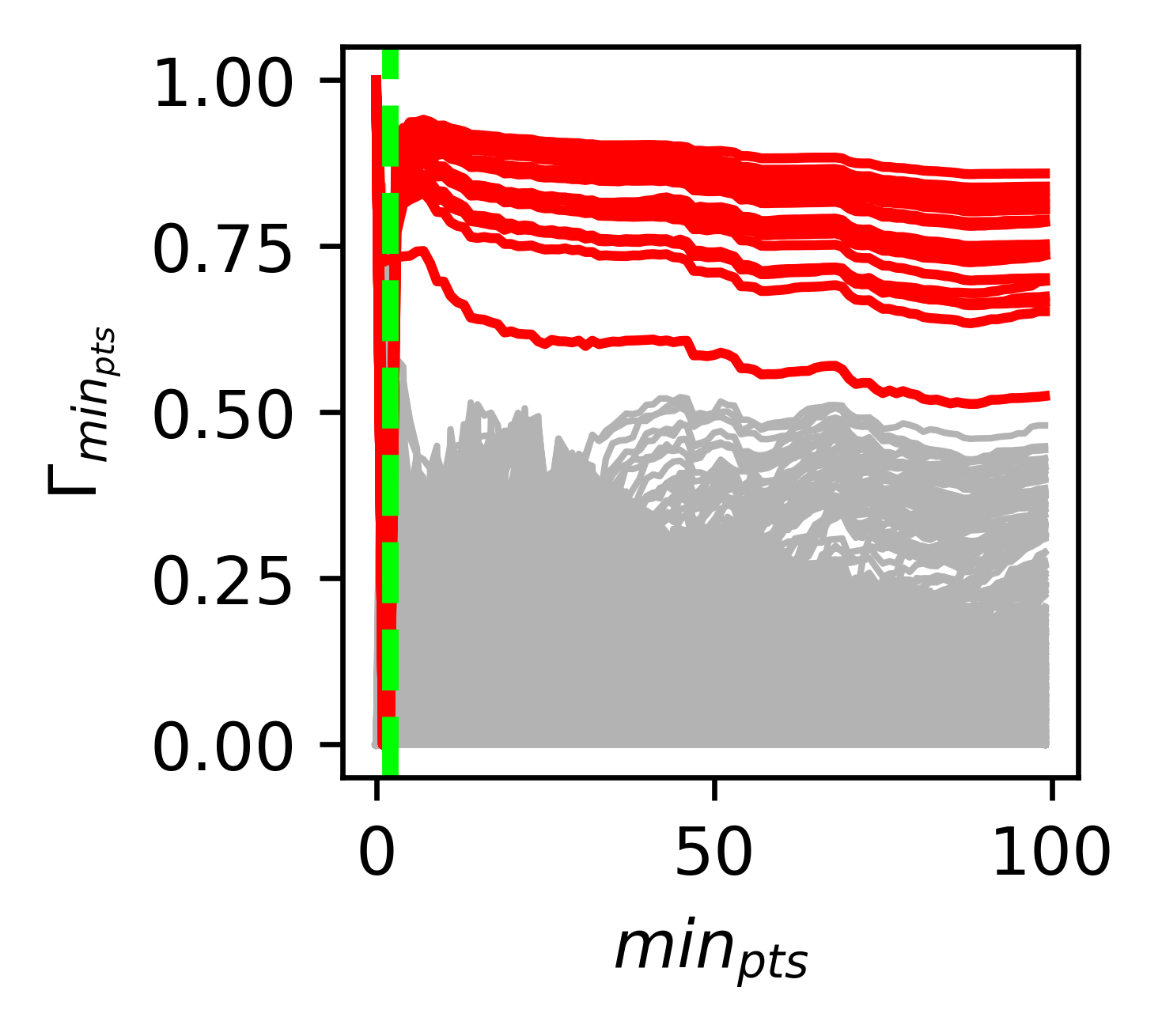}
\caption{Global-$P\Gamma$}
\label{fig:zenodo_banana_global_prof}
\end{subfigure}
\begin{subfigure}{2.3cm}
\centering\includegraphics[width=2.3cm]{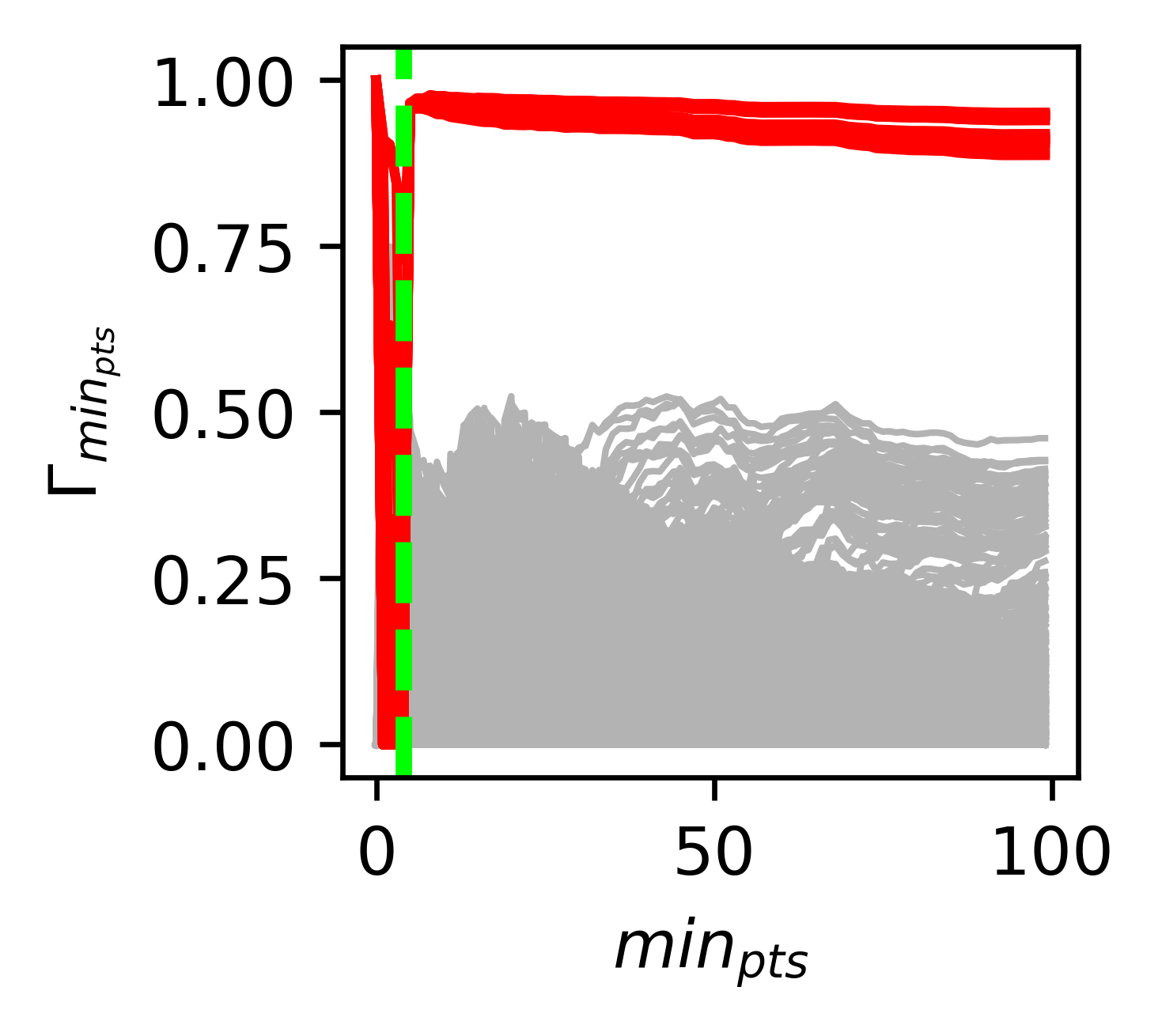}
\caption{Clumps-$P\Gamma$}
\label{fig:zenodo_banana_clump_prof}
\end{subfigure}
\begin{subfigure}{2.3cm}
\centering\includegraphics[width=2.3cm]{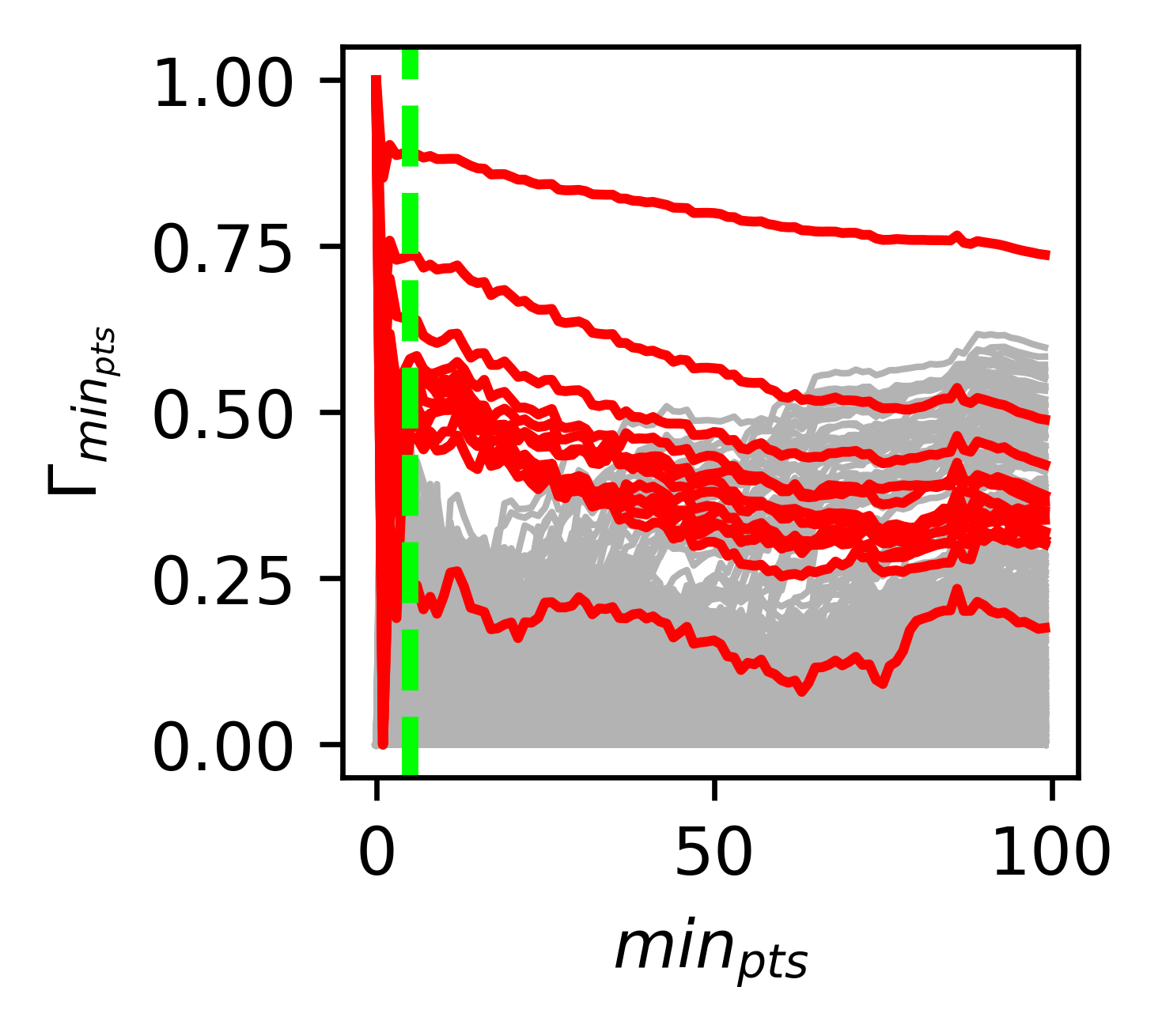}
\caption{Local-$P\Gamma$}
\label{fig:zenodo_banana_local_prof}
\end{subfigure}


\caption{Comparing GLOSH\textendash Profiles ($P\Gamma$) with Precision@n (P@n) obtained by GLOSH at different $min_{pts}$ values: The black line in Fig. \ref{fig:zenodo_banana_global_prec} to \ref{fig:zenodo_banana_local_prec} shows the P@n at every $min_{pts}$ value. The outliers profiles are in red while the inliers profiles are in gray. The green (dashed) line denotes the specific $min_{pts}$ value where GLOSH first achieves the best P@n within the range $[2,100]$ of $min_{pts}$ values.}

\label{fig:zenodo_datasets_pan_prof}
\raggedbottom
\end{figure}
We make the following key observations.
\textit{First}, Fig.~\ref{fig:zenodo_banana_global_prof} to \ref{fig:zenodo_banana_local_prof} show that for all the different kinds of outliers, the GLOSH scores in the outlier profiles fluctuate for low $min_{pts}$ values,
i.e., the GLOSH scores change significantly between consecutive $min_{pts}$ values and at different rates for different profiles. 
This happens when several outliers get lower GLOSH scores than inliers.
Therefore, in Fig.~\ref{fig:zenodo_banana_global_prec} to \ref{fig:zenodo_banana_local_prec}, GLOSH results in small P@n values for low $min_{pts}$. \textit{Second}, beyond a specific $min_{pts}$ value, the GLOSH scores in the profiles start undergoing minimal changes between consecutive $min_{pts}$ values. When they change, they do it at similar rates so that the resulting rankings tend to not change in a way that affects P@n.
\textit{Third}, for all kinds of outliers, we observe that the best P@n results correspond to the $min_{pts}$ values where the GLOSH scores start changing at a similar rate. However, there is an exception for local outliers, as
one cannot randomly choose a value beyond this specific $min_{pts}$ value and expect to achieve the best P@n in the presented experiment. 
Fig.~\ref{fig:zenodo_banana_local_prof} shows that local outlier profiles are more similar to inlier profiles when compared to the profiles of global outliers and clumps. There is no notable gap between the profiles of local outliers and inliers, as local outliers are close to the inliers.
Therefore, as the $min_{pts}$ value increases beyond the value where the GLOSH scores start changing at a similar rate,
some local outliers end up getting scores lower than several inliers. 
Overall, the experiments suggest that the $min_{pts}$ value where GLOSH scores start changing at a similar rate in the profiles can potentially yield the best results. Based on these insights, we develop a way to find this $min_{pts}$ value in the following.
\raggedbottom

\subsection{Dissimilarity between GLOSH score rankings for consecutive  $min_{pts}$ values}\label{sub_sec:stable_behavior_profiles}

Based on our findings in sub-section \ref{sub_sec:glosh_performance}, in this sub-section we study how 
GLOSH scores change between consecutive $min_{pts}$ values. One way to do that is by measuring the differences in the outlier rankings at consecutive $min_{pts}$ values. 
However, ordering of the data points w.r.t. GLOSH scores may be problematic. Two points in the same density region within a cluster may get identical GLOSH scores at consecutive $min_{pts}$ values. If one ranks the data points based on their GLOSH scores, any specific ordering of those two data points may differ between the consecutive $min_{pts}$ values. In such cases, the ordering of many inliers will dominate any measure of data point rank dissimilarity. 
  For outlier detection, we are primarily interested in outstanding score values or a small portion of the highest scores, as they often represent outliers. If points get identical GLOSH scores at consecutive $min_{pts}$ values, they should not reflect the difference between the rankings. 
  However, even though the ordering of points may differ, the scores of adjacent points in large stretches of a ranking may be very similar.
Therefore, instead of measuring the dissimilarity between sequences of ordered points, we measure the dissimilarity between the sorted sequences of GLOSH scores obtained at consecutive $min_{pts}$ values. 

   To measure the dissimilarity between the sorted GLOSH score sequences
   we use Pearson correlation, a popular symmetric metric that can measure the relationship between two outlier score sequences \cite{schubert2012evaluation}. We can express dissimilarity by subtracting the correlation from 1.
    For two sorted GLOSH score sequences obtained at $min_{pts} = k$ ($S_{k}$) and $min_{pts} = l$ ($S_{l}$), the Pearson dissimilarity $\Delta$ is defined as: 
    
\begin{equation}
\label{eq:pearson_dissim}
    \Delta(S_{k},S_{l}) = 1 - \begin{vmatrix}\frac{Cov(S_{k},S_{l})}{\sqrt{Var(S_{k})Var(S_{l})}}\end{vmatrix}
\end{equation}

\noindent where $Cov$ measures the covariance between sorted GLOSH score sequences for $min_{pts} = k$ and $min_{pts} = l$, and $Var$ measures the variance of the values in each of the sequences. 
When most GLOSH\textendash Profiles start changing at a similar rate, the resulting relative order and magnitude of GLOSH scores tend to undergo minimal changes between consecutive values of $min_{pts}$.
Therefore, the Pearson dissimilarity is unaffected by absolute value changes, as it focuses on the means' variations.
  However, for lower $min_{pts}$ values where the GLOSH scores in the outlier profiles may fluctuate, the covariance between the sorted GLOSH score sequences at consecutive $min_{pts}$ values should be high. Thus, we expect a high Pearson dissimilarity between the sorted GLOSH score sequences for low consecutive $min_{pts}$ values. Differently, for higher consecutive $min_{pts}$ values, the scores vary at a similar rate, and the Pearson dissimilarity tends to be 0.

\raggedbottom
 To study the behavior of GLOSH score rankings in more detail, we formally define the GLOSH Outlier Rank Dissimilarity\textendash Profile (ORD\textendash Profile) $R_{m_{max}}$. The ORD\textendash Profile quantifies the dissimilarity between the sorted sequences of GLOSH scores at consecutive $min_{pts}$ values as they increase.

\noindent \textbf{Definition 4.2. GLOSH Outlier Rank Dissimilarity\textendash Profile:} The GLOSH ORD\textendash Profile $R_{m_{max}}$ is an array of dissimilarity values for pairs of sorted sequences of GLOSH scores $S_{min_{pts}}$ obtained at consecutive $min_{pts}$ values in a range $[2,m_{max}]$:
\begin{align}
\label{eq:ord_prof}
    R_{m_{max}} &= \begin{bmatrix}
                       \Delta(S_{2},S_{3}) \\
                       \Delta(S_{3},S_{4}) \\
                       \vdots \\
                       \Delta(S_{m_{max}-1},S_{m_{max}})
                     \end{bmatrix}
    \end{align} 
\raggedbottom

\begin{figure}[t]
\centering

\begin{subfigure}{2.7cm}
\centering\includegraphics[width=2.7cm]{Figures/banana_global_prec.png}
\caption{Global-P@n}
\label{fig:zenodo_banana_global_prec2}
\end{subfigure}
\begin{subfigure}{2.7cm}
\centering\includegraphics[width=2.7cm]{Figures/banana_cluster_prec.png}
\caption{Clumps-P@n}
\label{fig:zenodo_banana_clump_prec2}
\end{subfigure}
\begin{subfigure}{2.7cm}
\centering\includegraphics[width=2.7cm]{Figures/banana_local_prec.png}
\caption{Local-P@n}
\label{fig:zenodo_banana_local_prec2}
\end{subfigure}

\begin{subfigure}{2.7cm}
\centering\includegraphics[width=2.7cm]{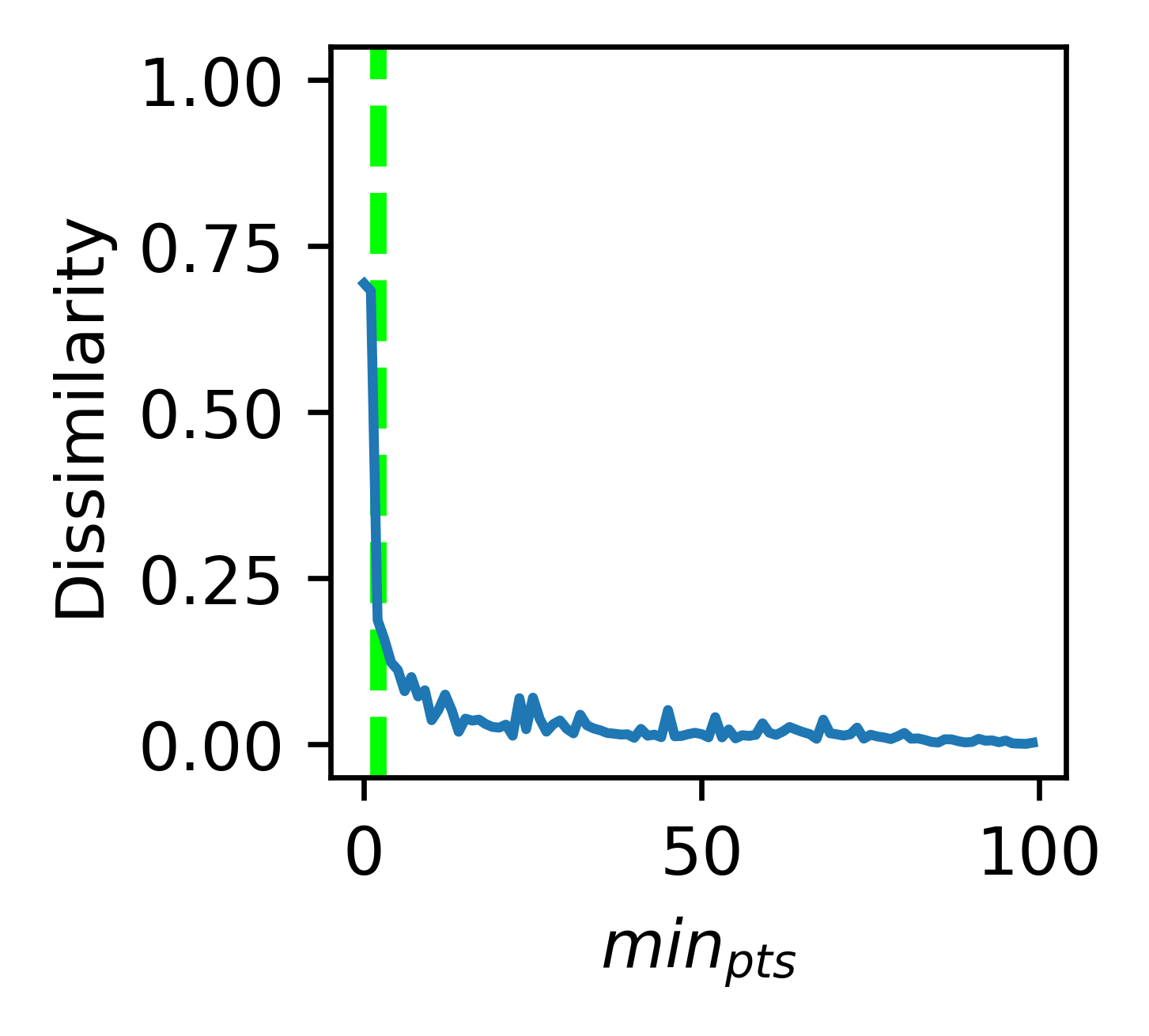}
\caption{Global-$R_{m_{max}}$}
\label{fig:zenodo_banana_global_dissim}
\end{subfigure}
\begin{subfigure}{2.7cm}
\centering\includegraphics[width=2.7cm]{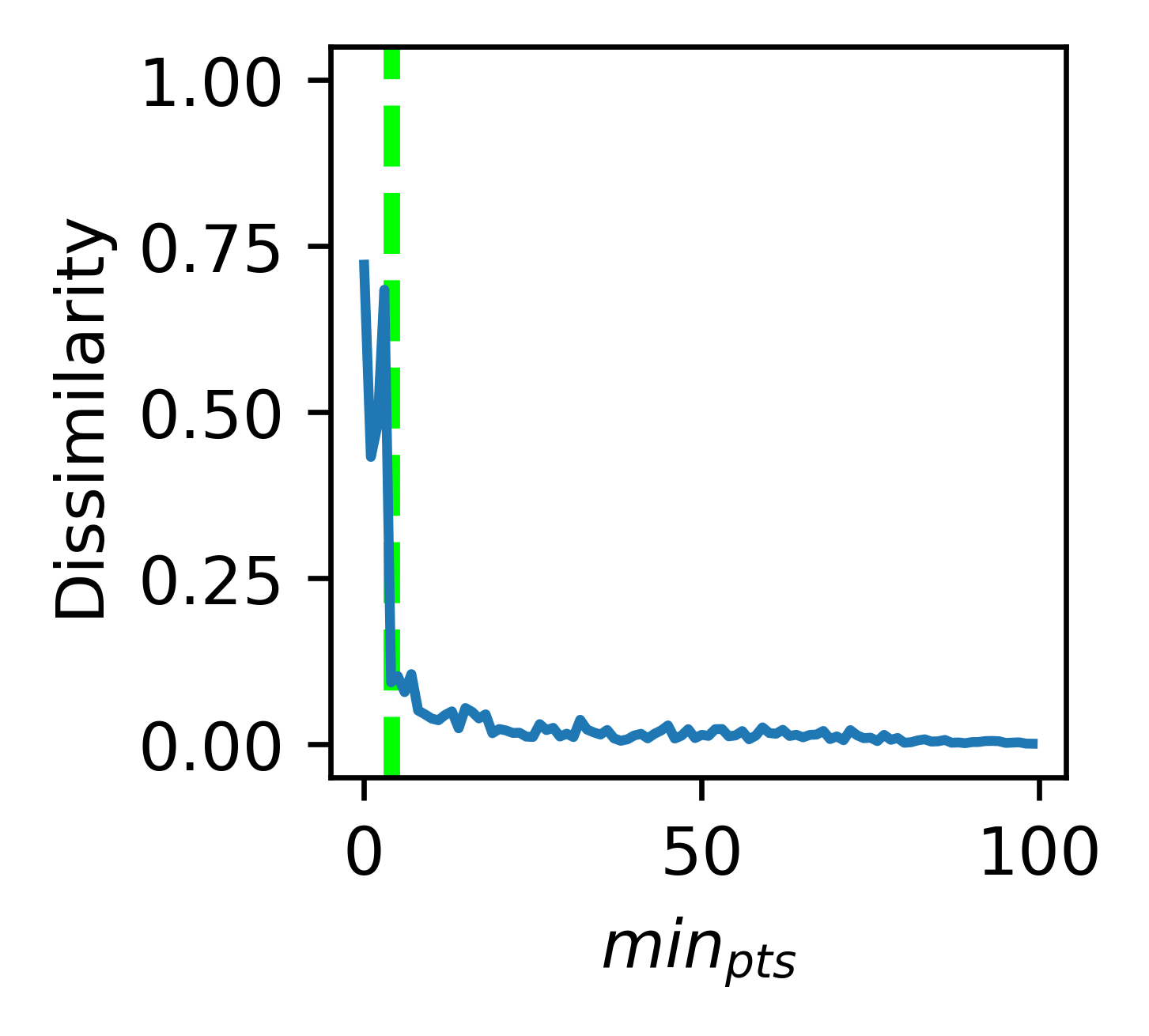}
\caption{Clumps-$R_{m_{max}}$}
\label{fig:zenodo_banana_clump_dissim}
\end{subfigure}
\begin{subfigure}{2.7cm}
\centering\includegraphics[width=2.7cm]{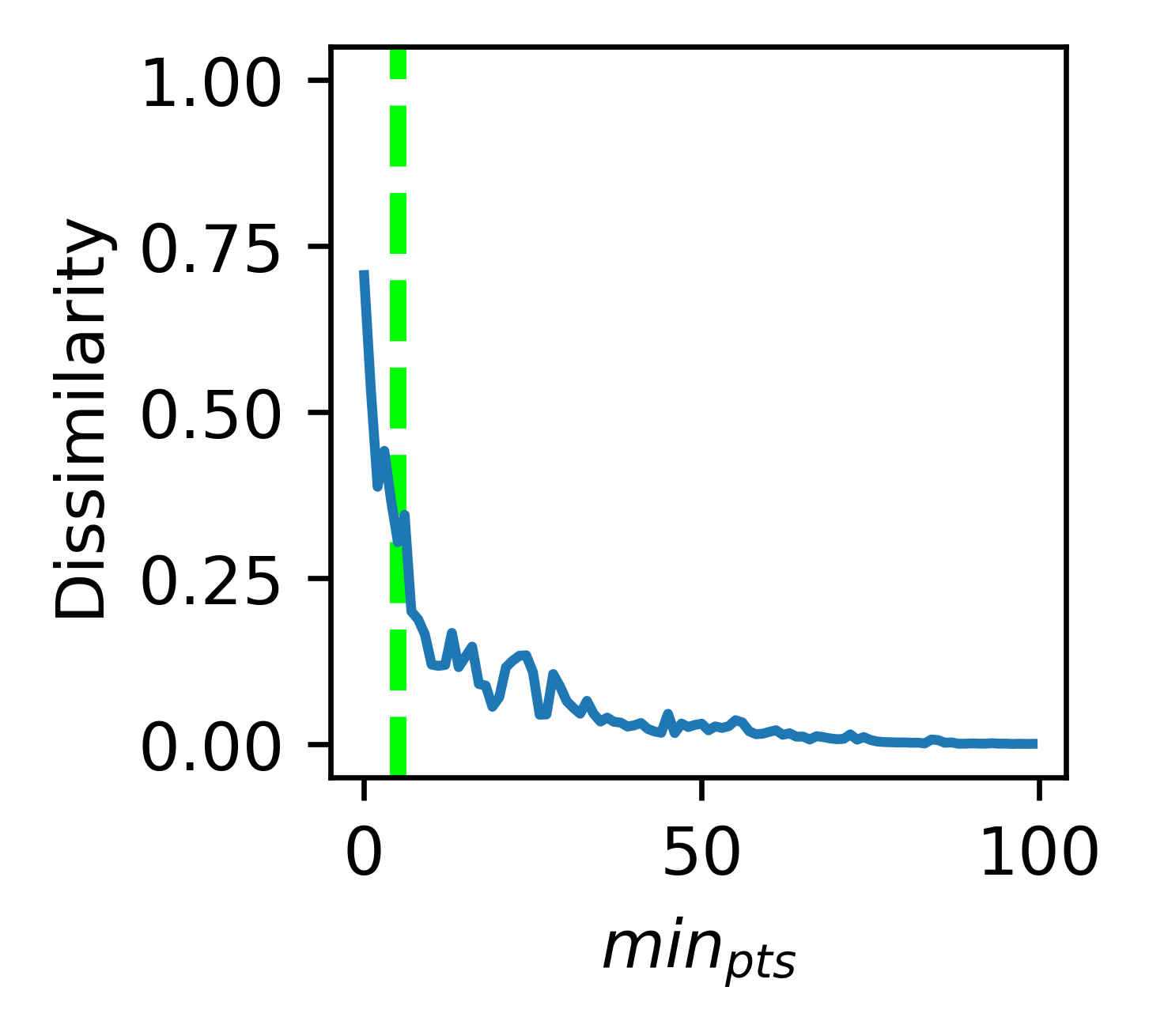}
\caption{Local-$R_{m_{max}}$}
\label{fig:zenodo_banana_local_dissim}
\end{subfigure}


\caption{Comparing the ORD\textendash Profiles ($R_{m_{max}}$) with GLOSH's P@n across different $min_{pts}$ values. We highlighted (with a dashed line)
the specific $min_{pts}$ value where GLOSH first records the best P@n within a
range [2, 100] of $min_{pts}$ values.}

\label{fig:zenodo_datasets_pan_ord}
\end{figure}

 As we define GLOSH\textendash Profiles from $min_{pts}=2$,
each dissimilarity score $r_i \in R_{m_{max}}$ is obtained 
between $min_{pts} = i+2$ and $min_{pts} = i+3$, where the indices $i$  starts from $0$.

Fig.~\ref{fig:zenodo_datasets_pan_ord} compares ORD\textendash Profiles with Precision@n (P@n) for a range of $min_{pts}$ values. \textit{First}, Fig.~\ref{fig:zenodo_banana_global_dissim} to \ref{fig:zenodo_banana_local_dissim} show
a high dissimilarity in the ORD\textendash Profile $R_{m_{max}}$ for lower $min_{pts}$ values, that evolve to 0 and form an ``elbow like'' structure.
The ``elbow'' is the point from where GLOSH scores start changing at a similar rate in most profiles. \textit{Second}, the $min_{pts}$ value where the ``elbow'' is formed corresponds to the $min_{pts}$ value where GLOSH achieves its best P@n. 
As local outliers are close to inliers, they do not get  GLOSH scores significantly higher than inliers. 
This happens as some local outliers end up getting lower GLOSH scores as inliers, as discussed in section \ref{sub_sec:glosh_performance}. For global outliers and clumps,  GLOSH scores are higher than inliers for a large range of $min_{pts}$ values.
In summary, Fig. \ref{fig:zenodo_datasets_pan_ord} suggests that choosing the $min_{pts}$ value directly at the ``elbow'' may potentially yield the best P@n.



\subsection{Finding the ``Best'' $min_{pts}$ value}\label{sub_sec:selective_profile_method}

Here, we propose Auto-GLOSH to find  the $m^{*}$ value, i.e., the $min_{pts}$ value at the 
``elbow'' 
in the ORD\textendash Profile and use it to yield the best 
results for GLOSH, as in sub-section \ref{sub_sec:stable_behavior_profiles}.


\noindent\textbf{Intuition.} Fig. \ref{sub_fig:autoGLOSH_intuition} illustrates the intuition behind Auto-GLOSH. Assuming a line AB is drawn connecting the maximum value (B) and end value (A) of an ORD\textendash Profile $R_{m_{max}}$, the latter with index $|R_{m_{max}}|$.
Perpendicular lines from AB to each value in the $R_{m_{max}}$ represent the ``shortest distance'' (orthogonal distance) between those values and AB. In this context, the value  at the ``elbow'' (E) is the one that has the largest orthogonal distance (shown as EO).

\noindent \textbf{Method.} Given an ORD\textendash Profile, $R_{m_{max}}$, find the maximum dissimilarity value in $R_{m_{max}}$ and call it $b$. 
Consider $i$ the index of $b$ in $R_{m_{max}}$,  which corresponds to the largest peak in the ORD\textendash Profile plot in Fig. \ref{fig:finding_elbow}. As shown in Fig. \ref{sub_fig:elbowa} and Fig. \ref{sub_fig:elbowb}, we compute a vector $\vv{AB}$ starting from the end value in $R_{m_{max}}$, $A = (|R_{m_{max}}|,R_{m_{max}}[|R_{m_{max}}|])$, and the maximum value, $B = (i,b)$, of the ORD\textendash Profile. Then, we compute $\vv{AB}$ as $B - A$. We will use $\vv{AB}$ as a tool to find the value in the ORD\textendash Profile that has the largest orthogonal distance to $\vv{AB}$.

\begin{figure}[t]
\centering
\begin{subfigure}{2.325cm}
\centering\includegraphics[width=2.325cm]{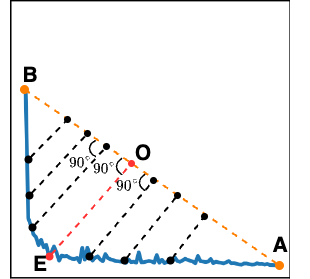}
\caption{Intuition}
\label{sub_fig:autoGLOSH_intuition}
\end{subfigure}
\begin{subfigure}{2.25cm}
\centering\includegraphics[width=2.25cm]{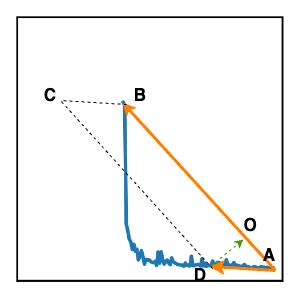}
\caption{Parallelogram}
\label{sub_fig:elbowa}
\end{subfigure}
\begin{subfigure}{2.25cm}
\centering\includegraphics[width=2.25cm]{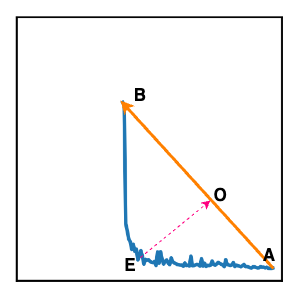}
\caption{Max Distance}
\label{sub_fig:elbowb}
\end{subfigure}
\caption{Illustrating the process of finding the Elbow of the Outlier Rank Dissimilarity\textendash Profile on the Banana Dataset with Global Outliers}
\label{fig:finding_elbow}
\end{figure}

Next, we compute the orthogonal distances between each dissimilarity value lying between A and B in the ORD\textendash Profile, and $\vv{AB}$. To do that, consider the parallelogram $ABCD$ as in Fig. \ref{sub_fig:elbowa} where $D$ is a value between $A$ and $B$ in ORD\textendash Profile. Assuming $\vv{AB}$ as the base of the parallelogram $ABCD$, the orthogonal distance between $D$ and $\vv{AB}$, represented as $\vv{DO}$, is the height of $ABCD$. Our goal is to find the $||\vv{DO}||$, where $||.||$ is the magnitude. In classical geometry, the area of a parallelogram is computed as $base \times height$, which is $||\vv{AB}|| \times ||\vv{DO}||$ w.r.t. Fig. \ref{sub_fig:elbowa}.
As shown in \cite{anton2013elementary,lay2003linear}, using vectors, the area of the parallelogram $ABCD$ is:
\begin{equation}\label{eq:orthogonal_dist}
\begin{split}
    ||\vv{AD} \times \vv{AB}|| &= ||\vv{AB}|| \times ||\vv{DO}|| \implies ||\vv{DO}|| = \frac{||\vv{AD} \times \vv{AB}||}{||\vv{AB}||}
\end{split}
\end{equation}
\raggedbottom
\noindent where, $\vv{AD}$ is computed as $D-A$.
We compute the orthogonal distances of each dissimilarity value lying between $A$ and $B$ in the ORD\textendash Profile and find the value with the largest distance as the elbow point.
This can be done efficiently by continuously tracking the largest distance as we compute the orthogonal distances of each dissimilarity value.
Assuming $r_i$ as the value at the elbow (discussed in sub-section \ref{sub_sec:stable_behavior_profiles}), we can find $r_i$ at $min_{pts} = i+3$, on measuring the dissimilarity between the sequences at $min_{pts} = i+3$ and $min_{pts} = i+2$. We take $i+3$ as the value of $m^{*}$. 

\subsection{Complexity Analysis}
\label{sub_sec:complexity_analysis}
Our method extracts the GLOSH\textendash Profiles from different HDBSCAN* hierarchies at different $min_{pts}$ values. The naïve way to build these profiles is to run HDBSCAN* for every $min_{pts}$ value up to $m_{max}$. This may be inefficient and require $\mathcal{O}(n^2m_{max})$ time as a single HDBSCAN* run takes $\mathcal{O}(n^2)$, assuming $n$ is the dataset size and $m_{max}<<n$ ($m_{max}$ is typically chosen to be below $100$ in practice).
However, \emph{CORE-SG}  permits a more efficient process \cite{neto2022core}. With a single HDBSCAN* run at $min_{pts}=m_{max}$, the  \emph{CORE-SG}$_{m_{max}}$  can be build from 
 $MST_{m_{max}}$ and $m_{max}$-\emph{Nearest Neighbor Graph} \cite{neto2022core}. 
By replacing the complete graph with the \emph{CORE-SG}$_{m_{max}}$, all $MST_{min_{pts}}$ with $min_{pts}$ values up to $m_{max}$ can be obtained with an asymptotic computational complexity of $\mathcal{O}(nm_{max}^2\log n)$, which is lower than a single run of HDBSCAN* for $m_{max}<<n$.

As shown in \cite{campello2015hierarchical}, the GLOSH scores for all the datapoints can be computed in $\mathcal{O}(n)$ time using the $MST_{min_{pts}}$ and equation \ref{eq:glosh}.
From the scores, the GLOSH\textendash Profiles can be built in $\mathcal{O}(nm_{max})$ time.
From the profiles, the Pearson dissimilarity is measured between consecutive GLOSH score sequences. Expressed as $\Delta(.,.)$ in equation \ref{eq:pearson_dissim}, calculating the dissimilarity requires $\mathcal{O}(n)$ time, as measuring $Cov$ and $Var$ requires iterating over each GLOSH score in the sequences. Therefore, building the ORD\textendash Profile $R_{m_{max}}$ needs $\mathcal{O}(nm_{max})$ time. Then, assessing the $m^*$ value has $\mathcal{O}(n)$ time, as finding the maximum value in $R_{m_{max}}$ and obtaining the largest orthogonal distance are procedures with linear cost in relation to $n$.

In summary, the overall asymptotic computational cost of our method is  upper bounded by the first run of HDBSCAN* to build the \emph{CORE-SG}$_{m_{max}}$, which is $\mathcal{O}(n^2)$. 
Note that efficient and parallel algorithms for $MST$ construction and cluster extraction in HDBSCAN* exist, which can further accelerate our method. For instance, \cite{gu2022parallel} shows that HDBSCAN*'s complexity can be reduced to $\mathcal{O}(n\log^2n)$ in sequential time and $\mathcal{O}(\log^2n\log\log n)$ in parallel time. Another approach based on data summarization and parallelism performed well for large datasets \cite{santos2021}.

\label{tex:best_minpts}

\section{Unsupervised Labelling of
Potential Outliers}
\label{sec:threshold_estimation}

Here, we address the problem of choosing a threshold for labelling inliers and
potential outliers. We first study the distribution of the GLOSH scores at the
$m^*$ (\ref{sub_sec:stable_gloshScore_distribution}), then we design POLAR to find the threshold (\ref{sub_sec:inlier_potentialOutlier_separation}).

\subsection{GLOSH scores at the $m^*$}
\label{sub_sec:stable_gloshScore_distribution}

In practice, the number of outliers is usually unknown, which may be challenging to estimate.
We believe that a pattern or gap in the sorted sequence of GLOSH scores at the $m^{*}$ value can indicate 
a suitable threshold for this value. 
A ``gap'' would be a noticeable deviation or abrupt increase in the sorted sequence, suggesting a potential decrease in the density of points and indicating the possibility of outliers.

\begin{figure}[t]
\centering

\begin{subfigure}{2.5cm}
\centering\includegraphics[width=2.5cm]{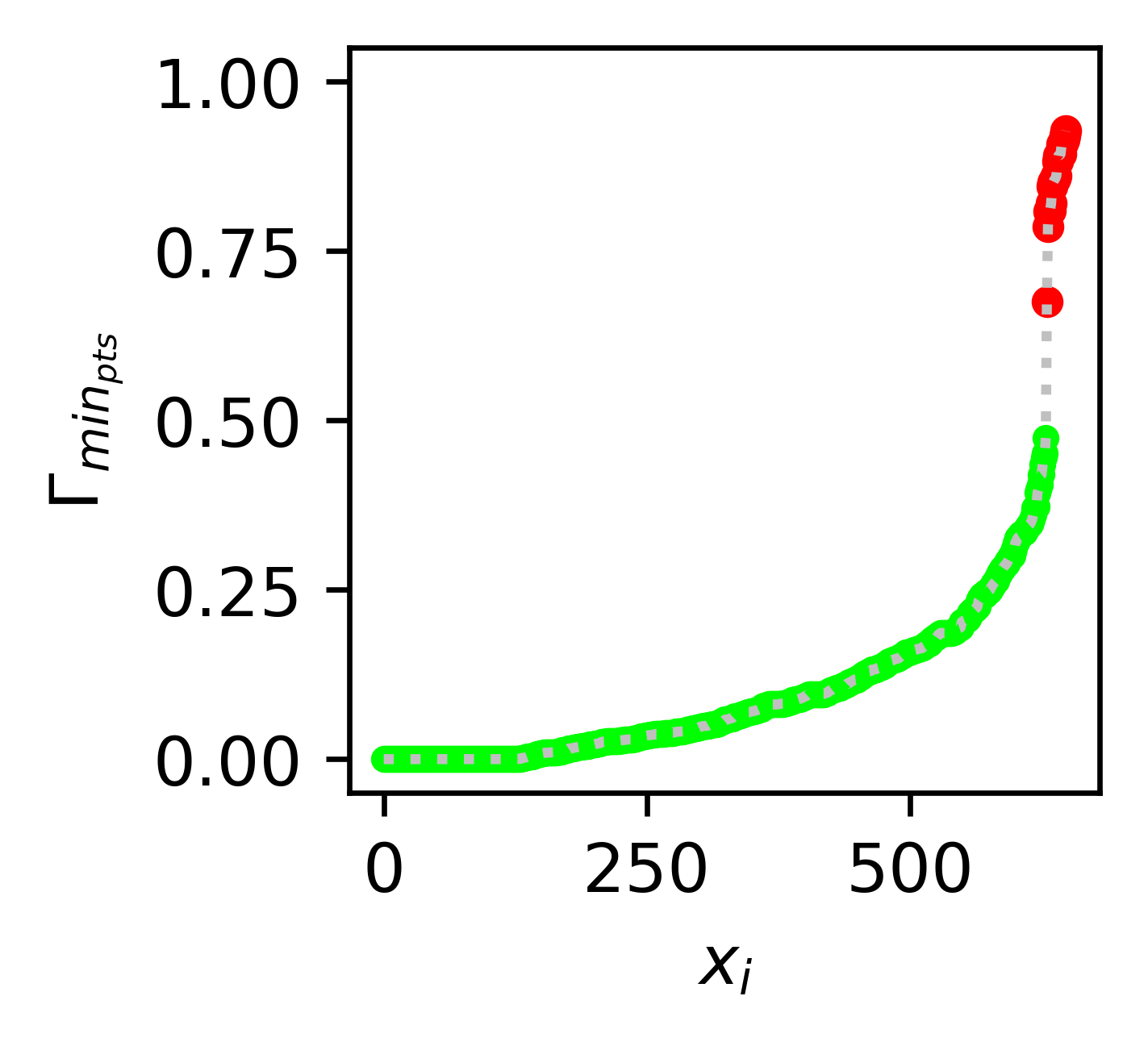}
\caption{Global}
\label{fig:StablegloshScore_distribution_banana-global}
\end{subfigure}
\begin{subfigure}{2.5cm}
\centering\includegraphics[width=2.5cm]{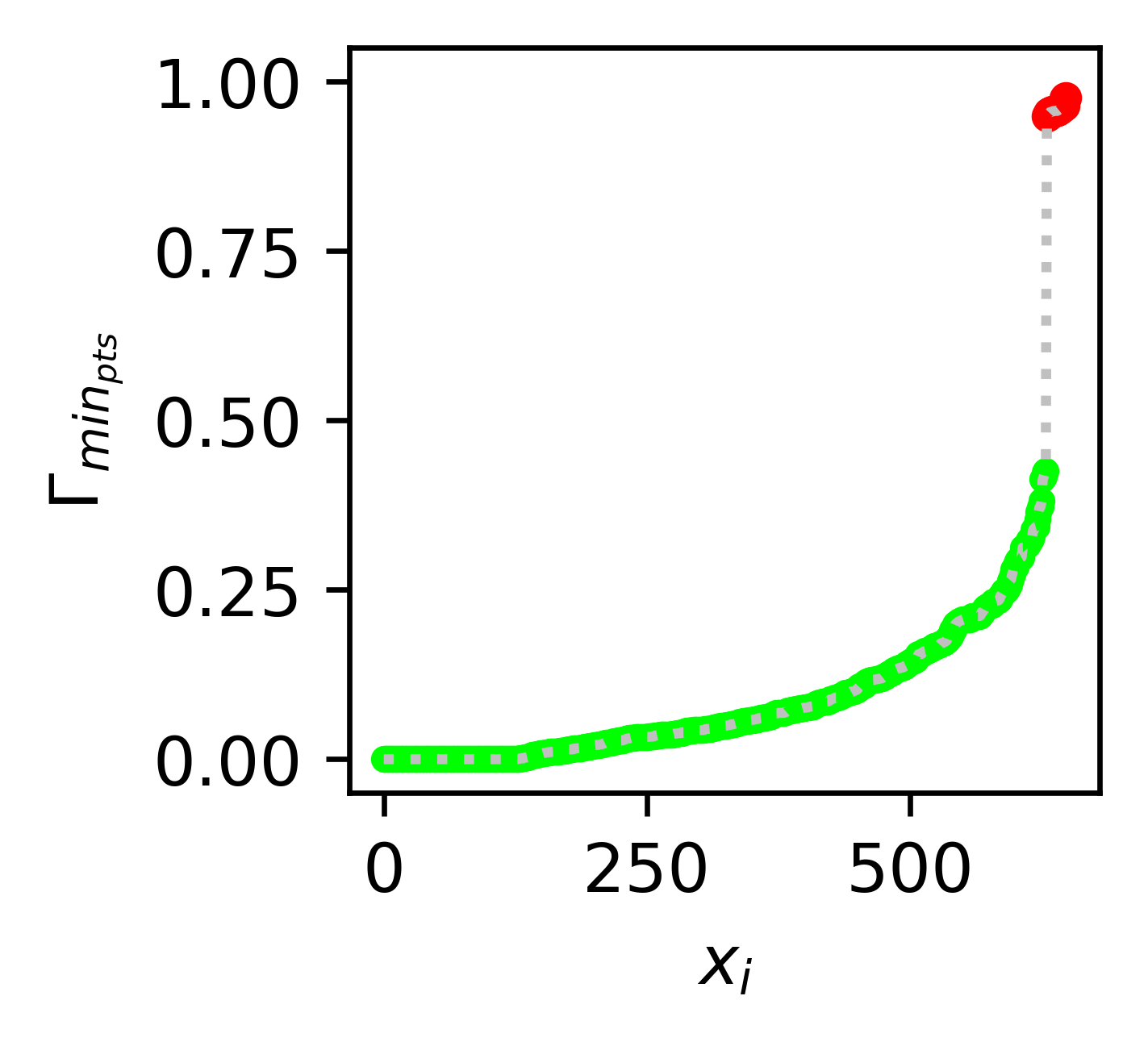}
\caption{Clumps}
\label{fig:StablegloshScore_distribution_banana-clump}
\end{subfigure}
\begin{subfigure}{2.5cm}
\centering\includegraphics[width=2.5cm]{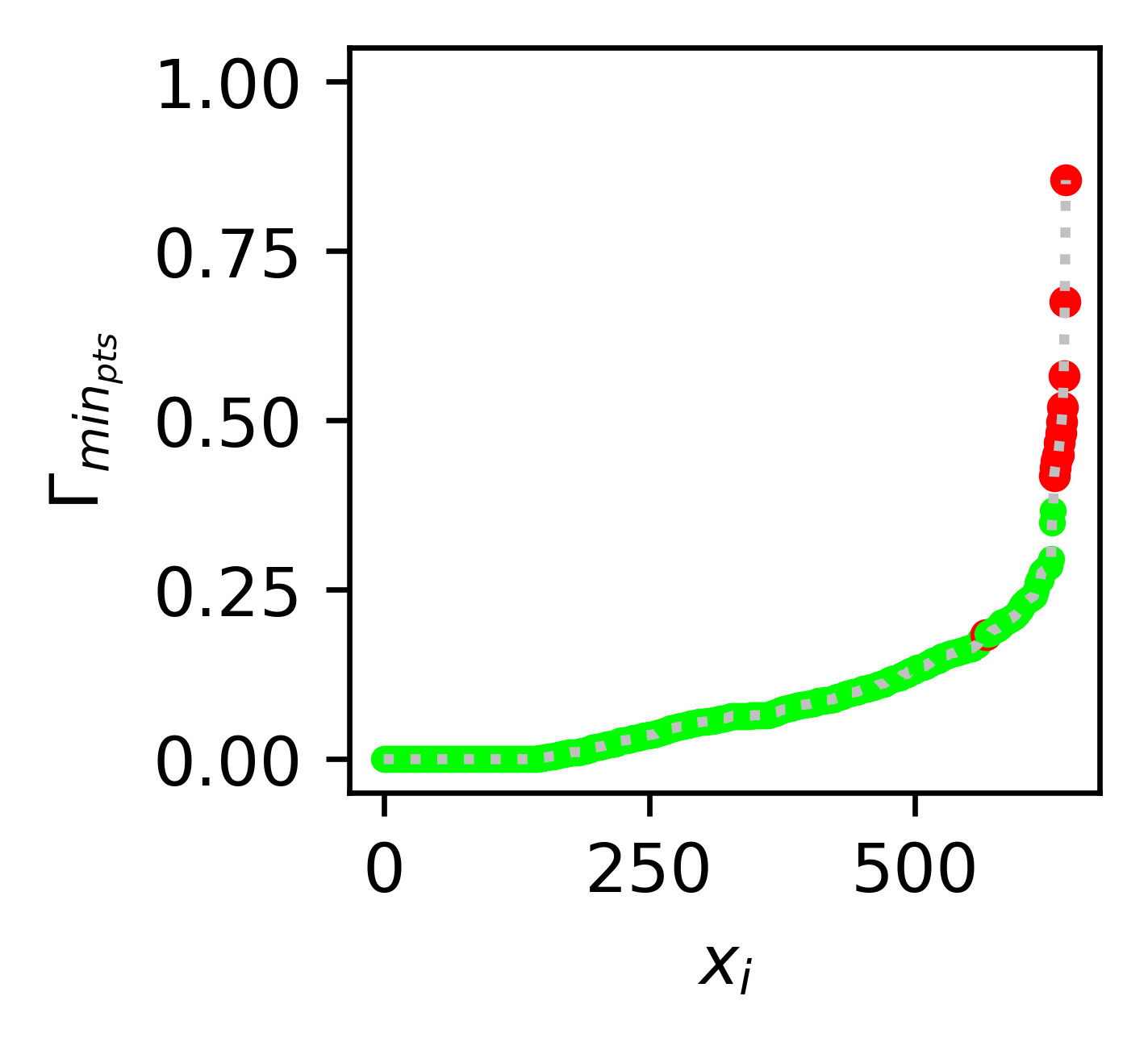}
\caption{Local}
\label{fig:StablegloshScore_distribution_banana-local}
\end{subfigure}
\\

\caption{Banana Dataset: Sorted Sequence of GLOSH scores at $min_{pts} = m^{*}$. The x-axis presents the data points $x_i$ in ascending order of their GLOSH scores, and the y-axis represents the corresponding GLOSH scores $\Gamma_{m^*}(x_i)$. The green dots represent the GLOSH scores of inliers, and the red represents that of outliers.}
\label{fig:StablegloshScore_distribution_zenodo}
\end{figure}

The example in  Fig. \ref{fig:StablegloshScore_distribution_zenodo} illustrates the intuition behind the threshold estimation we propose.  As expected, outliers receive high GLOSH scores and are predominantly located at the end of the sequence for the assessed $m^*$ value. 
One can see that the sorted GLOSH score sequences form a "knee" when the scores start deviating from the initial linear trend and show accelerated growth.
This is true for all kinds of outliers in Fig. \ref{fig:StablegloshScore_distribution_banana-global} to \ref{fig:StablegloshScore_distribution_banana-local}. 
Such a ``knee'' is formed by GLOSH scores of points that significantly deviate from the majority of points and may serve as a suitable threshold to label potential outliers.
However, as shown in Fig. \ref{fig:StablegloshScore_distribution_zenodo}, there may be a few points with high deviations and GLOSH scores beyond the ``knee'' that behave like outliers but are labeled as inliers. 
To avoid classifying these points as outliers, it is necessary to adjust the selection of the threshold ``beyond the knee'' of the sorted sequence. Based on these insights, we present our threshold estimation approach in the following sub-section.

\subsection{Automatically Labelling Potential Outliers}
\label{sub_sec:inlier_potentialOutlier_separation}

In this sub-section, we design POLAR, a fully unsupervised approach to label potential outliers by using the sorted GLOSH score sequences obtained at $m^{*}$ value. 
In the first step, our approach finds the ``knee'' in the GLOSH score sequences at the $m^{*}$ value. However, to avoid incorrectly labeling the inliers that may lie beyond the ``knee'', an adjustment strategy is necessary for choosing the threshold.


\noindent \textbf{Finding the ``knee'' GLOSH score:} To find the ``knee'', we use a similar strategy to the one previously presented in section \ref{sub_sec:selective_profile_method} to locate the ``elbow'' in the ORD\textendash Profile. As illustrated in Fig. \ref{fig:knee-finding_bananaClump}, 
we firstly compute the vector $\vv{AB}$, as the difference between the last GLOSH score $B$ and the first GLOSH score $A$ in the sorted sequence. Next, we compute the orthogonal distances between each GLOSH score and $\vv{AB}$. The GLOSH score in the sequence with the maximum orthogonal distance is identified as the ``knee'' GLOSH score. 

\begin{figure}[t]
\centering
\begin{subfigure}{5.7cm}
\centering\includegraphics[width=5.7cm]{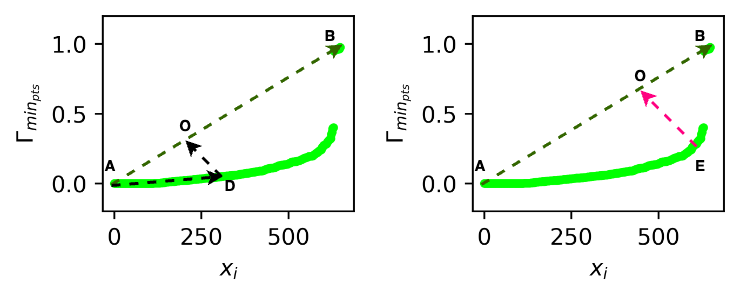}
\caption{}
\label{fig:knee-finding_bananaClump}
\end{subfigure}
\begin{subfigure}{5.7cm}
\centering\includegraphics[width=5.7cm]{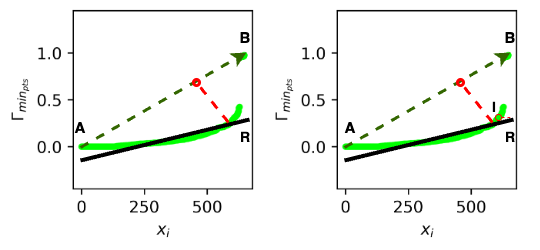}
\raggedbottom
\caption{}
\label{fig:banana_regression}
\end{subfigure}


\caption{Banana Dataset with clumps: (a) Finding the ``knee'' point in the sorted sequence of GLOSH scores at $min_{pts} = m^*$. The ``knee'' GLOSH score E has the maximum orthogonal distance to $\protect\vv{AB}$. (b) Adjusted threshold estimation at $m^*$.}

\label{fig:finding-thres_bananaClump}

\end{figure}
\raggedbottom
\noindent \textbf{Adjusting the Inlier Threshold:} To adjust a threshold beyond the ``knee'' GLOSH score in the sorted sequence, we first aim to capture the progression of the inlier GLOSH scores in the sorted sequence. We use a simple linear regression model \cite{james2023linear} to estimate the trend of the inlier GLOSH scores in the sequence. The idea here is to capture the ``almost'' linear trend of the GLOSH scores observed at the beginning of the sorted sequence for inliers. The linear regression model estimates a quantitative value $\hat{Y}$ based on a predictor variable $X$, assuming there is approximately a linear relationship between X and Y. The linear regression model can be represented as:
\begin{equation}
\label{eq:linear_regression_general}
    \hat{Y} = \beta_0 + \beta_1\times X
\end{equation}
For a dataset with $N$ points, the indexes of each GLOSH score in the sorted sequence can be represented as $[x_1,x_2,\ldots, x_N]$. We use the GLOSH scores before the ``knee'' GLOSH score in the sorted sequence to estimate the model coefficients $\beta_0$ and $\beta_1$. The optimal values for $\beta_0$ and $\beta_1$ are estimated by minimizing the Mean Squared Error (MSE) between the estimated $\hat{Y}$ and the known $Y$. In Fig. \ref{fig:banana_regression}, the estimated regression line is represented as a black solid line.
Using the regression model, we estimate the highest GLOSH score $R$ at index $x_N$. 
Intuitively, this estimated GLOSH score $R$ reflects the score that could have been reached if there were only inliers in the dataset and the scores in the sequence followed the trend that was observed up to the ``knee''.
Then, we search the GLOSH score, $I$, that is the most similar to $R$ in the sorted sequence between the ``knee'' and $x_N$ and choose it as the adjusted threshold. 

\label{tex:best_threshold}

\section{Experimental Analysis}
\label{sec:exp_analysis}

\subsection{Dataset}
\label{subsec:dataset}

To evaluate Auto-GLOSH and POLAR, we use a total of 69 datasets.
This includes the Banana dataset from sub-section \ref{sub_sec:glosh_performance}, along with two additional datasets, Anisotropic and Circular, obtained from \cite{koncar_2018_1171077}. Different types of outliers (global, local, clumps) are generated on these datasets, following the approach described in sub-section \ref{sub_sec:glosh_performance}. Additionally, we make semi-synthetic datasets by using the
real inliers from real one class classification (OCC) datasets and generating synthetic outliers on them. To do this, we use fifteen different real OCC datasets \cite{han2022adbench} that are popularly used in the literature:
MVTec-AD\_zipper, HEPATITIS, LETTER, PIMA, STAMPS, VERTEBRAL, VOWELS, WDBC, WINE, WPBC, YEAST, BREASTW, CARDIO, CARDIOTOCOGRAPHY, and 20news\_3. They are across various domains: image, healthcare, documents, biology, chemistry, and NLP.

\noindent \textbf{Why synthetic outliers?} \textit{Firstly}, nonsynthetic datasets commonly do not have their ``real outliers'' (as statistically defined in the literature \cite{hawkins1980identification}) previously labeled and available for analysis.
For this reason, it is usual to rely on real OCC datasets, making them popular in outlier detection studies. These are originally classification datasets where one class is downsampled to fit into an outlier detection setting, assuming that the downsampled class (re-labeled as ``outlier class'') has the traditional characteristics of ``real outliers''. The characteristics of these ``labeled outliers'' (data points in the ``outlier class'')  are not fully understood in the literature \cite{steinbuss2021benchmarking}.
    This provides limited scope for testing our methods on different types of outliers as the specific outliers types that are present in the so-called ``outlier class'' are often unknown. We aim to evaluate our methods across different outlier types, which is not possible using the real OCC datasets.


\textit{Secondly}, as also discussed in \cite{steinbuss2021benchmarking}, in many real OCC datasets the ``labeled outliers'' may often deviate from the characteristics typically associated with outliers in the outlier detection literature (global, local, clumps). To support this argument, we report in Table \ref{tab:realData_results} the best Precision@n (P@n) obtained on real OCC datasets of state-of-the-art neighborhood-based methods, GLOSH, KNN, and LOF, and parameter-free methods, ABOD \cite{kriegel2008angle} and COPOD \cite{li2020copod}.
The overall results in Table \ref{tab:realData_results} reflect that across most of the datasets, none of the methods work well in identifying the ``labeled outliers'', even with their best parameters. 
This indicates that many of the ``labeled outliers'' do not have the properties typically associated with the way these state-of-the-art methods define outliers.
Additionally, we illustrate the behavior of the ``labeled outliers'' through GLOSH\textendash Profiles to allow us to understand the properties of these data points better. Fig. \ref{fig:glosh-profile_realdatasets} shows that the profiles of many ``labeled outliers'' achieve GLOSH scores lower than the ``labeled inliers'', as such points sometimes behave like inlier profiles. To derive meaningful conclusions in our investigations, it is essential to have outliers that conform to their statistical definition \cite{hawkins1980identification}, i.e., they exhibit characterizable deviations from the inliers.



\begin{table}[]
\centering
\resizebox{!}{2.0cm}{%
\begin{tabular}{|c|ccccc|}
\hline
\textbf{Dataset} & \textbf{GLOSH} & \textbf{KNN} & \textbf{LOF} & \textbf{ABOD} & \textbf{COPOD} \\ \hline
MVTec-AD\_zipper & .62 & .62 & .6 & .6 & .58 \\ \hline
HEPATITIS & .3 & .3 & .3 & .23 & .46 \\ \hline
LETTER & .3 & .39 & .54 & .1 & .04 \\ \hline
PIMA & .55 & .55 & .51 & .55 & .48 \\ \hline
STAMPS & .25 & .25 & .19 & .16 & .41 \\ \hline
VERTEBRAL & .03 & .03 & .1 & .03 & 0 \\ \hline
VOWELS & .54 & .54 & .38 & .34 & 0 \\ \hline
WDBC & .5 & .5 & .6 & .4 & .8 \\ \hline
WINE & .4 & .4 & .4 & .1 & .4 \\ \hline
WPBC & .19 & .19 & .21 & .19 & .21 \\ \hline
YEAST & .28 & .27 & .32 & .26 & .25 \\ \hline
BREASTW & .94 & .94 & .32 & .95 & .94 \\ \hline
CARDIO & .53 & .53 & .28 & .44 & .52 \\ \hline
CARDIOTOCOGRAPHY & .41 & .4 & .35 & .38 & .36 \\ \hline
20news\_3 & .13 & .13 & .2 & .13 & .13 \\ \hline
\end{tabular}}
\caption{Results on real one-class classification datasets. For GLOSH, KNN, and LOF, we report the best results obtained across the commonly used $min_{pts}$ and $k$ values in the literature and practice \cite{campello2015hierarchical,jayasinghe2019milky,jarry2020aircraft,marques2022similarity,yin2023station}.}
\label{tab:realData_results}
\end{table}

\begin{figure}[t]
\centering
\begin{subfigure}{2.1cm}
\centering\includegraphics[trim={0 0 2.9cm 0},clip,width=2.1cm]{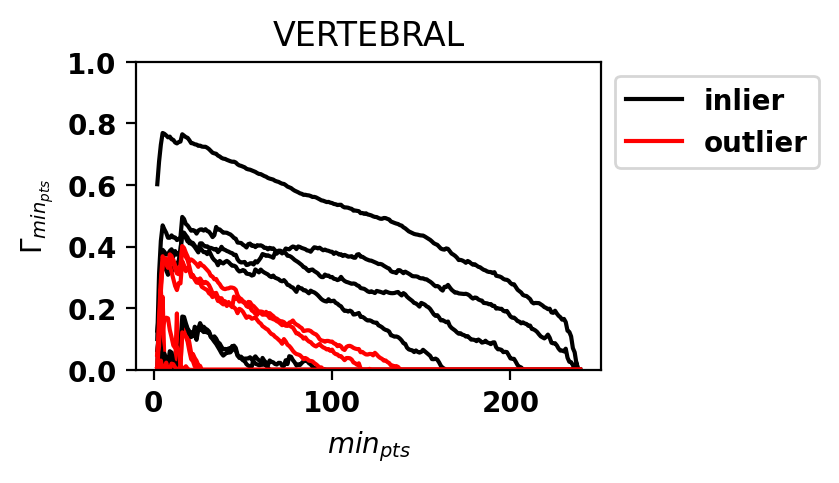}
\end{subfigure}
\begin{subfigure}{2.1cm}
\centering\includegraphics[trim={0 0 2.9cm 0},clip,width=2.1cm]{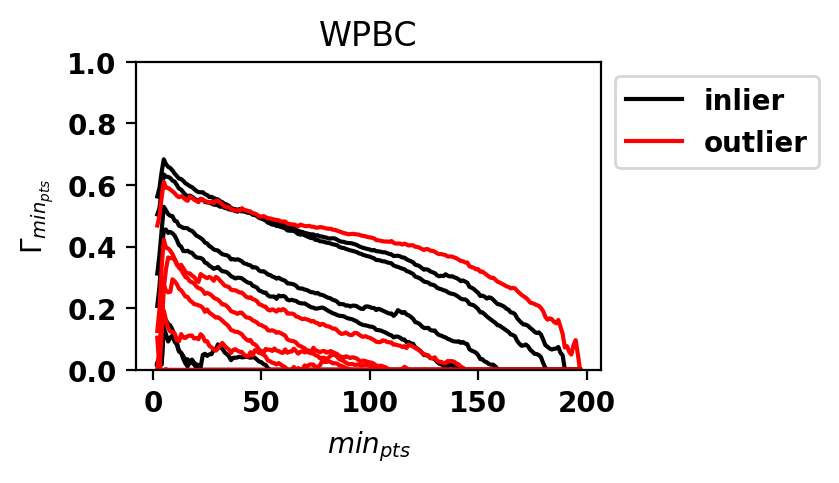}
\end{subfigure}
\begin{subfigure}{2.1cm}
\centering\includegraphics[trim={0cm 0 2.9cm 0},clip,width=2.1cm]{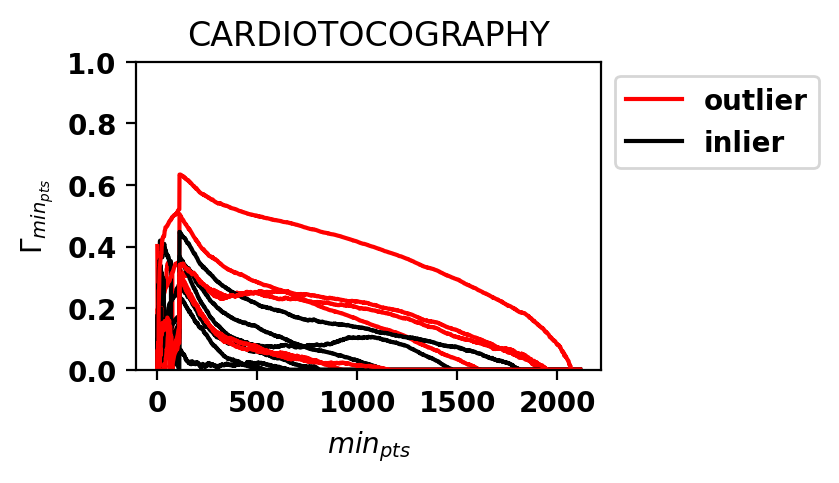}
\end{subfigure}
\begin{subfigure}{2.1cm}
\centering\includegraphics[trim={0 0 2.9cm 0},clip,width=2.1cm]{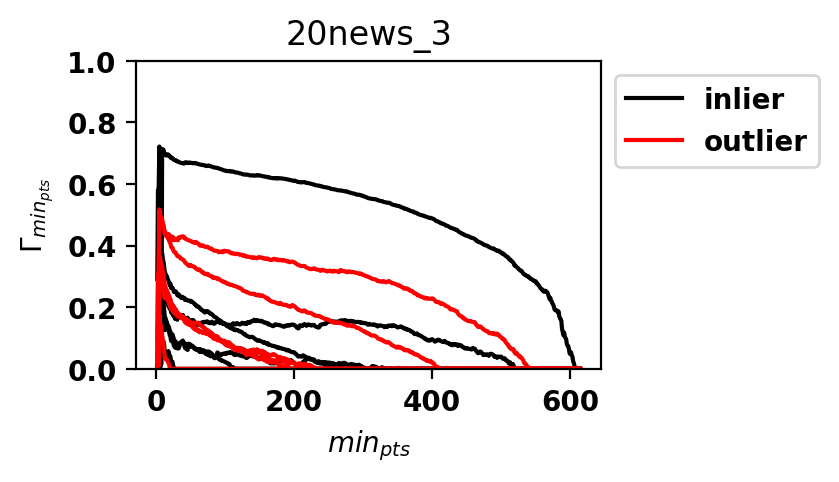}
\end{subfigure}

\caption{GLOSH\textendash Profiles of real one-class classification datasets \cite{han2022adbench}: VERTEBRAL, WPBC, CARDIOTOCOGRAPHY, and 20news\_3. To have a clear view of the individual profiles, we randomly choose six inlier and outlier class data points and plot their profiles. The red profiles represent the outlier class and the black represents the inlier class.}
\label{fig:glosh-profile_realdatasets}
\end{figure}

Due to the aforementioned reasons, we generated synthetic outliers following the approach described in sub-section \ref{sub_sec:glosh_performance} to evaluate Auto-GLOSH and POLAR.
This approach to generating outliers is more meaningful than downsampling a class, as a generative model is used to create outliers with specific characteristics close to those of ``real outliers'', as statistically defined in the literature.

\subsection{Auto-GLOSH}
\label{sub_sec:result_auto_GLOSH}

\begin{table*}[t]
\centering
\resizebox{\textwidth}{2.0cm}{%
\begin{tabular}{|c|cccc|cccccccccccccccccccccccc|}
\hline
\multirow{3}{*}{\textbf{Dataset}} & \multicolumn{4}{c|}{\multirow{2}{*}{\textbf{Auto-GLOSH}}} & \multicolumn{24}{c|}{\textbf{GLOSH}} \\ \cline{6-29} 
 & \multicolumn{4}{c|}{} & \multicolumn{4}{c|}{\pmb{$\Gamma_{5}$}} & \multicolumn{4}{c|}{\pmb{$\Gamma_{10}$}} & \multicolumn{4}{c|}{\pmb{$\Gamma_{25}$}} & \multicolumn{4}{c|}{\pmb{$\Gamma_{50}$}} & \multicolumn{4}{c|}{\pmb{$\Gamma_{100}$}} & \multicolumn{4}{c|}{\textbf{*Best}} \\ \cline{2-29} 
 & \textbf{G} & \textbf{C} & \textbf{L} & \textbf{M} & \textbf{G} & \textbf{C} & \textbf{L} & \multicolumn{1}{c|}{\textbf{M}} & \textbf{G} & \textbf{C} & \textbf{L} & \multicolumn{1}{c|}{\textbf{M}} & \textbf{G} & \textbf{C} & \textbf{L} & \multicolumn{1}{c|}{\textbf{M}} & \textbf{G} & \textbf{C} & \textbf{L} & \multicolumn{1}{c|}{\textbf{M}} & \textbf{G} & \textbf{C} & \textbf{L} & \multicolumn{1}{c|}{\textbf{M}} & \textbf{G} & \textbf{C} & \textbf{L} & \textbf{M} \\ \hline
MVTec-AD\_zipper & 1 & 1 & 1 & 1 & 0 & 0 & 1 & \multicolumn{1}{c|}{.56} & 0 & 0 & 1 & \multicolumn{1}{c|}{0} & 1 & 1 & 1 & \multicolumn{1}{c|}{1} & 1 & 1 & 1 & \multicolumn{1}{c|}{1} & 1 & 1 & 1 & \multicolumn{1}{c|}{1} & 1 & 1 & 1 & 1 \\
HEPATITIS & 1 & 1 & 1 & 1 & 1 & 1 & 1 & \multicolumn{1}{c|}{1} & 1 & 1 & 1 & \multicolumn{1}{c|}{1.0} & 1 & 1 & 1 & \multicolumn{1}{c|}{1} & 1 & 1 & 1 & \multicolumn{1}{c|}{1} & - & - & - & \multicolumn{1}{c|}{-} & 1 & 1 & 1 & 1 \\
LETTER & 1 & 1 & .9 & 1 & 1 & .13 & .93 & \multicolumn{1}{c|}{.25} & 1 & .5 & .83 & \multicolumn{1}{c|}{.09} & 1 & .29 & .81 & \multicolumn{1}{c|}{.15} & 1 & 0 & .9 & \multicolumn{1}{c|}{.04} & 1 & 1 & .9 & \multicolumn{1}{c|}{1} & 1 & 1 & .93 & 1 \\
PIMA & 1 & 1 & .86 & 1 & 1 & 0 & .72 & \multicolumn{1}{c|}{.42} & 1 & .04 & .86 & \multicolumn{1}{c|}{.14} & 1 & 1 & .68 & \multicolumn{1}{c|}{1} & 1 & 1 & .63 & \multicolumn{1}{c|}{1} & 1 & 1 & .59 & \multicolumn{1}{c|}{1} & 1 & 1 & .86 & 1 \\
STAMPS & 1 & 1 & .35 & 1 & 1 & 0 & .5 & \multicolumn{1}{c|}{0} & 1 & 0 & .35 & \multicolumn{1}{c|}{0} & 1 & 1 & .35 & \multicolumn{1}{c|}{1} & .93 & 1 & .35 & \multicolumn{1}{c|}{1} & .93 & 1 & .35 & \multicolumn{1}{c|}{1} & 1 & 1 & .57 & 1 \\
VERTEBRAL & .9 & 1 & .7 & .95 & .9 & .2 & .7 & \multicolumn{1}{c|}{.04} & .9 & 1 & .7 & \multicolumn{1}{c|}{.95} & .9 & 1 & .7 & \multicolumn{1}{c|}{.95} & .9 & 1 & .7 & \multicolumn{1}{c|}{.95} & .9 & 1 & .7 & \multicolumn{1}{c|}{.95} & .9 & 1 & .7 & .95 \\
VOWELS & 1 & 1 & .88 & 1 & 1 & .14 & .92 & \multicolumn{1}{c|}{.38} & 1 & .02 & .88 & \multicolumn{1}{c|}{.51} & 1 & 1 & .88 & \multicolumn{1}{c|}{1} & 1 & 1 & .84 & \multicolumn{1}{c|}{1} & 1 & 1 & .88 & \multicolumn{1}{c|}{1} & 1 & 1 & .92 & 1 \\
WDBC & 1 & 1 & .66 & .96 & 1 & .16 & .66 & \multicolumn{1}{c|}{.59} & 1 & 0 & .66 & \multicolumn{1}{c|}{.07} & 1 & 1 & .66 & \multicolumn{1}{c|}{.96} & 1 & 1 & .66 & \multicolumn{1}{c|}{.96} & 1 & 1 & .66 & \multicolumn{1}{c|}{.96} & 1 & 1 & 1 & .96 \\
WINE & 1 & 1 & 1 & 1 & 1 & 0 & 1 & \multicolumn{1}{c|}{0} & 1 & 1 & 1 & \multicolumn{1}{c|}{1} & 1 & 1 & 1 & \multicolumn{1}{c|}{1} & 1 & 1 & 1 & \multicolumn{1}{c|}{1} & 1 & 1 & 1 & \multicolumn{1}{c|}{1} & 1 & 1 & 1 & 1 \\
WPBC & 1 & 1 & .8 & 1 & 1 & 0 & .8 & \multicolumn{1}{c|}{0} & 1 & 1 & .8 & \multicolumn{1}{c|}{1} & 1 & 1 & 1 & \multicolumn{1}{c|}{1} & 1 & 1 & .8 & \multicolumn{1}{c|}{1} & 1 & 1 & 1 & \multicolumn{1}{c|}{1} & 1 & 1 & 1 & 1 \\
YEAST & 1 & .42 & .87 & .76 & .04 & .04 & .89 & \multicolumn{1}{c|}{.76} & .73 & .44 & .87 & \multicolumn{1}{c|}{.79} & 1 & .42 & .46 & \multicolumn{1}{c|}{.82} & 1 & .4 & .46 & \multicolumn{1}{c|}{.82} & 1 & .4 & .44 & \multicolumn{1}{c|}{.81} & 1 & .46 & .89 & .83 \\
BREASTW & 1 & 1 & .22 & 1 & .9 & 0 & 0 & \multicolumn{1}{c|}{.22} & 1 & 0 & .09 & \multicolumn{1}{c|}{.08} & 1 & 1 & .22 & \multicolumn{1}{c|}{1} & 1 & 1 & .22 & \multicolumn{1}{c|}{1} & 1 & 1 & .22 & \multicolumn{1}{c|}{1} & 1 & 1 & .22 & 1 \\
CARDIO & 1 & .79 & .87 & .92 & 1 & .14 & .56 & \multicolumn{1}{c|}{.43} & 1 & .15 & .87 & \multicolumn{1}{c|}{.42} & 1 & .85 & .8 & \multicolumn{1}{c|}{.94} & 1 & .84 & .79 & \multicolumn{1}{c|}{.94} & 1 & .79 & .79 & \multicolumn{1}{c|}{.91} & 1 & .85 & .89 & .94 \\
CARDIOTOCOGRAPHY & 1 & 1 & .89 & 1 & 1 & .13 & .76 & \multicolumn{1}{c|}{.14} & 1 & .1 & .91 & \multicolumn{1}{c|}{.01} & 1 & .01 & .89 & \multicolumn{1}{c|}{0} & 1 & 0 & .89 & \multicolumn{1}{c|}{.005} & 1 & 1 & .86 & \multicolumn{1}{c|}{1} & 1 & 1 & .93 & 1 \\
20news\_3 & 1 & 1 & .95 & .98 & 0 & 0 & 0 & \multicolumn{1}{c|}{.08} & 0 & 0 & .95 & \multicolumn{1}{c|}{.25} & 0 & 0 & .95 & \multicolumn{1}{c|}{.08} & 1 & 1 & .95 & \multicolumn{1}{c|}{.98} & 1 & 1 & .91 & \multicolumn{1}{c|}{.98} & 1 & 1 & .95 & 1 \\ \hline
Anisotropic & 1 & 1 & .69 & - & .15 & .26 & .61 & \multicolumn{1}{c|}{-} & 1 & 0 & .61 & \multicolumn{1}{c|}{-} & 1 & 1 & .61 & \multicolumn{1}{c|}{-} & 1 & 1 & .46 & \multicolumn{1}{c|}{-} & 1 & 1 & .53 & \multicolumn{1}{c|}{-} & 1 & 1 & .76 & - \\
Banana & 1 & 1 & .92 & - & 1 & .05 & .69 & \multicolumn{1}{c|}{-} & 1 & 1 & .92 & \multicolumn{1}{c|}{-} & 1 & 1 & .92 & \multicolumn{1}{c|}{-} & 1 & 1 & .38 & \multicolumn{1}{c|}{-} & 1 & 1 & .07 & \multicolumn{1}{c|}{-} & 1 & 1 & .92 & - \\
Circular & 1 & 1 & .76 & - & .94 & 1 & .3 & \multicolumn{1}{c|}{-} & 1 & 1 & .46 & \multicolumn{1}{c|}{-} & 1 & 1 & .76 & \multicolumn{1}{c|}{-} & 1 & 1 & .76 & \multicolumn{1}{c|}{-} & 1 & 1 & .76 & \multicolumn{1}{c|}{-} & 1 & 1 & .76 & - \\ \hline
\end{tabular}}
\caption{Auto-GLOSH vs GLOSH: Precision@n obtained with global outliers (G), clumps (C), local outliers (L), and mixed outliers (M)}
\label{tab:autoGLOSH_vs_glosh}
\end{table*}


\noindent \textbf{Setup.} For evaluating Auto-GLOSH, we focus on three key questions: (I) Does the Auto-GLOSH estimated $m^{*}$ value match the best GLOSH performance? (II) Can the estimated $m^{*}$ value surpass or match results obtained with commonly used $min_{pts}$ values in GLOSH? (III) Can GLOSH with the $m^{*}$ value outperform existing state-of-the-art outlier detection methods? A comparative analysis is conducted between Auto-GLOSH and GLOSH with some commonly used $min_{pts}$ values in the literature and practice. 
Performance is assessed using Precision@n (P@n). 
We also compare the performance at the $m^{*}$ value with that achieved at the best $min_{pts}$ value within range $[2, 100]$, i.e., the value where GLOSH obtains its highest P@n.
Furthermore, we evaluate the performance at the $m^{*}$ value against the state-of-the-art outlier detection methods such as LOF and KNN, which operate w.r.t. a neighborhood parameter $k$.
Specifically, we compare the results obtained using Auto-GLOSH to those of LOF and KNN across commonly used $k$ values. We also compare our results against parameter-free methods such as ABOD and COPOD and present a comparative analysis of the runtime required by our method.

\noindent \textbf{Auto-GLOSH Vs. GLOSH.} In Table \ref{tab:autoGLOSH_vs_glosh},
for most datasets, our Auto-GLOSH estimated $m^{*}$ value yields the best P@n that is achievable using GLOSH (represented as *Best). 
Comparatively, using the commonly used $min_{pts}$ values ($\Gamma_5$ to $\Gamma_{100}$) may result in a low P@n if the $min_{pts}$ value is not chosen properly.
Notably, for the Anisotropic dataset with local outliers, the Auto-GLOSH gets a higher P@n than all the commonly used $min_{pts}$ values. 
For some datasets with local outliers, such as CARDIO or CARDIOTOCOGRAPHY, Auto-GLOSH did not match the best P@n; however, it is close to the best. This shows that Auto-GLOSH is able to find a $min_{pts}$ value that can yield the best or nearly the best GLOSH results.



\noindent \textbf{Auto-GLOSH Vs. Neighborhood-based Methods.}  In Table \ref{tab:autoGLOSH_vs_knn_lof}, both KNN and LOF can outperform the best results achievable using GLOSH for some of the $k$ values, especially for local outliers and clumps.
Auto-GLOSH is built upon GLOSH and designed to extract the best results achievable with GLOSH. Consequently, it is expected that Auto-GLOSH may not surpass the performance of KNN and LOF when GLOSH itself does not outperform them. However, the overall results show that KNN and LOF only perform well for the best $k$ values, i.e.,  $k$ must be chosen carefully. In LETTER with local outliers, increasing $k$ decreases P@n for KNN but keeps it stable for LOF. For PIMA with local outliers, P@n decreases for both KNN and LOF as $k$ increases. Examples like these make it unclear which $k$ value to choose when the underlying data distribution is unknown. This emphasizes the necessity for Auto-GLOSH, which finds the $min_{pts}$ value capable of achieving high P@n on most datasets.

\raggedbottom
\begin{table*}[]
\centering
\resizebox{\textwidth}{!}{%
\begin{tabular}{|c|cccc|cccc|cccc|cccc|cccc|cccc|cccc|}
\hline
\multirow{3}{*}{\textbf{Dataset}} & \multicolumn{4}{c|}{\multirow{2}{*}{\textbf{Auto-GLOSH}}} & \multicolumn{4}{c|}{\pmb{$k=5$}} & \multicolumn{4}{c|}{\pmb{$k=10$}} & \multicolumn{4}{c|}{\pmb{$k=25$}} & \multicolumn{4}{c|}{\pmb{k=50}} & \multicolumn{4}{c|}{\pmb{k=100}} & \multicolumn{4}{c|}{\multirow{2}{*}{\textbf{*Best GLOSH}}} \\ \cline{6-25}
 & \multicolumn{4}{c|}{} & \multicolumn{4}{c|}{\textbf{KNN / LOF}} & \multicolumn{4}{c|}{\textbf{KNN / LOF}} & \multicolumn{4}{c|}{\textbf{KNN / LOF}} & \multicolumn{4}{c|}{\textbf{KNN / LOF}} & \multicolumn{4}{c|}{\textbf{KNN / LOF}} & \multicolumn{4}{c|}{} \\ \cline{2-29} 
 & \textbf{G} & \textbf{C} & \textbf{L} & \textbf{M} & \textbf{G} & \textbf{C} & \textbf{L} & \textbf{M} & \textbf{G} & \textbf{C} & \textbf{L} & \textbf{M} & \textbf{G} & \textbf{C} & \textbf{L} & \textbf{M} & \textbf{G} & \textbf{C} & \textbf{L} & \textbf{M} & \textbf{G} & \textbf{C} & \textbf{L} & \textbf{M} & \textbf{G} & \textbf{C} & \textbf{L} & \textbf{M} \\ \hline
MVTec-AD\_zipper & 1 & 1 & 1 & 1 & 1 / 0 & .48 / .07 & 1 / 1 & .86 / .26 & 1 / 0 & .48 / .03 & 1 / 1 & 1 / 0 & 1 / 0 & .81 / 0 & 1 / 1 & 1 / 1 & 1 / 1 & 1 / 1 & 1 / 1 & 1 / 1 & 1 / 1 & 1 / 1 & 1 / 1 & 1 / 1 & 1 & 1 & 1 & 1 \\
HEPATITIS & 1 & 1 & 1 & 1 & 1 / .42 & 1 / 0 & 1 / 0 & 1 / .5 & 1 / 1 & 1 / 1 & 1 / .75 & 1 / 1 & 1 / 1 & 1 / 1 & 1 / 1 & 1 / 1 & 1 / 1 & 1 / 1 & 1 / 1 & 1 / 1 & - / - & - / - & - / - & 1 / 1 & 1 & 1 & 1 & 1 \\
LETTER & 1 & 1 & .9 & 1 & 1 / .35 & .61 / .1 & .95 / .95 & .9 / .43 & 1 / .52 & .66 / .06 & .95 / 1 & .96 / .44 & 1 / .82 & .93 / .17 & .9 / .98 & 1 / .34 & 1 / .94 & 1 / 0 & .9 / .98 & 1 / .29 & 1 / .98 & 1 / .02 & .9 / .98 & 1 / .13 & 1 & 1 & .93 & 1 \\
PIMA & 1 & 1 & .86 & 1 & 1 / .42 & .56 / .12 & .86 / .9 & .94 / .29 & 1 / .58 & .6 / .04 & .86 / .95 & 1 / .25 & 1 / .84 & .98 / 0 & .63 / .9 & 1 / .2 & 1 / .98 & 1 / 0 & .63 / .9 & 1 / .44 & 1 / .98 & 1 / 1 & .59 / .68 & 1 / 1 & 1 & 1 & .86 & 1 \\
STAMPS & 1 & 1 & .35 & 1 & 1 / .25 & .41 / .29 & .5 / .78 & .93 / .18 & 1 / .51 & .64 / .22 & .35 / .64 & .96 / .12 & 1 / .74 & .9 / .09 & .35 / .5 & 1 / .15 & .96 / .96 & 1 / .93 & .35 / .5 & 1 / .84 & .96 / 1 & 1 / 1 & .35 / .42 & 1 / 1 & 1 & 1 & .57 & 1 \\
VERTEBRAL & .9 & 1 & .7 & .95 & .95 / .23 & .33 / .23 & .71 / .64 & .95 / .23 & .95 / .47 & .66 / .23 & .78 / .78 & .95 / .19 & .95 / .9 & 1 / .95 & .78 / .78 & .95 / .95 & .95 / .95 & 1 / .95 & .78 / .78 & .95 / .95 & .95 / .95 & 1 / 1 & .78 / .78 & .95 / .95 & .9 & 1 & .7 & .95 \\
VOWELS & 1 & 1 & .88 & 1 & 1 / .82 & .63 / .17 & .92 / .92 & .95 / .55 & 1 / .97 & .68 / .09 & .88 / .92 & .99 / .81 & 1 / 1 & .9 / .43 & .88 / 1 & 1 / 1 & 1 / 1 & 1 / 1 & .84 / 1 & 1 / 1 & 1 / 1 & 1 / 1 & .88 / 1 & 1 / 1 & 1 & 1 & .92 & 1 \\
WDBC & 1 & 1 & .66 & .96 & 1 / .08 & .36 / .11 & .66 / 1 & .77 / .67 & 1 / .19 & .38 / .25 & .66 / 1 & .96 / .25 & 1 / .52 & .8 / .25 & .66 / 1 & .96 / .07 & 1 / .94 & 1 / .94 & .66 / 1 & .96 / .96 & 1 / 1 & 1 / 1 & .66 / 1 & .96 / 1 & 1 & 1 & 1 & .96 \\
WINE & 1 & 1 & 1 & 1 & 1 / .08 & .58 / .08 & 1 / 1 & 1 / 0 & 1 / .16 & .91 / 0 & 1 / 1 & 1 / .8 & 1 / 1 & 1 / 1 & 1 / 1 & 1 / 1 & 1 / 1 & 1 / 1 & 1 / 1 & 1 / 1 & 1 / 1 & 1 / 1 & 1 / 1 & 1 / 1 & 1 & 1 & 1 & 1 \\
WPBC & 1 & 1 & .8 & 1 & 1 / 1 & .4 / 0 & .8 / .4 & 1 / 0 & 1 / 1 & .66 / 0 & .8 / .8 & 1 / 0 & 1 / 1 & 1 / .93 & 1 / .8 & 1 / 1 & 1 / 1 & 1 / 1 & .8 / 1 & 1 / 1 & 1 / 1 & 1 / 1 & 1 / 1 & 1 / 1 & 1 & 1 & 1 & 1 \\
YEAST & 1 & .42 & .87 & .76 & 1 / .27 & .48 / .15 & .8 / .63 & .76 / .35 & 1 / .29 & .47 / .17 & .76 / .78 & .79 / .47 & 1 / .36 & .54 / .26 & .7 / .73 & .82 / .61 & 1 / .6 & .55 / .44 & .7 / .72 & .82 / .67 & 1 / .78 & .57 / .51 & .69 / .71 & .81 / .75 & 1 & .46 & .89 & .83 \\
BREASTW & 1 & 1 & .22 & 1 & 1 / 0 & .31 / 0 & .27 / .09 & .58 / 0 & 1 / .02 & .38 / 0 & .29 / .13 & .72 / 0 & 1 / 0 & .7 / 0 & .38 / .29 & .97 / 0 & 1 / .63 & 1 / .04 & .38 / .5 & 1 / .06 & 1 / .9 & 1 / .68 & .34 / .36 & 1 / .87 & 1 & 1 & .22 & 1 \\
CARDIO & 1 & .79 & .87 & .92 & 1 / .83 & .62 / .06 & .94 / .81 & .86 / .39 & 1 / .89 & .63 / .07 & .89 /.93 & .87 / .43 & 1 / .98 & .75 / .25 & .82 / .98 & .94 / .77 & 1 / .99 & .86 / .5 & .81 / .96 & .94 / .92 & 1 / 1 & .85 / .85 & .8 / .93 & .91 / .95 & 1 & .85 & .89 & .94 \\
CARDIOTOCOGRAPHY & 1 & 1 & .89 & 1 & 1 / .02 & .58 / .07 & .93 / .71 & .86 / .33 & 1 / .07 & .58 / .05 & .91 / .89 & .91 / .39 & 1 / .18 & .68 / .18 & .9 / .95 & .98 / .33 & 1 / .35 & .87 / .22 & .9 / .96 & .99 / .13 & 1 / .67 & .98 / .01 & .87 / .95 & 1 / .08 & 1 & 1 & .93 & 1 \\
20news\_3 & 1 & 1 & .95 & .98 & 1 / 0 & .46 / .01 & .95 / .73 & .88 / .41 & 1 / 0 & .39 / .03 & .95 / .95 & .83 / .61 & 1 / 0 & .53 / .06 & .95 / .95 & .98 / .26 & 1 / 0 & .74 /.01 & .95 / .95 & .98 / .1 & 1 / .98 & 1 / 1 & .91 / .95 & .98 / 1 & 1 & 1 & .95 & 1 \\ \hline
Anisotropic & 1 & 1 & .69 & - & 1 / .15 & .89 / 0 & .53 / .46 & - & 1 / .26 & 1 / 0 & .53 / .61 & - & 1 / 1 & 1 / 1 & .53 / .84 & - & 1 / 1 & 1 / 1 & .53 / .53 & - & 1 / 1 & 1 / 1 & .53 / .53 & - & 1 & 1 & .76 & - \\
Banana & 1 & 1 & .92 & - & 1 / .42 & 1 / .21 & 1 / .72 & - & 1 / 1 & 1 / .31 & 1 / 1 & - & 1 / 1 & 1 / 1 & 1 / 1 & - & 1 / 1 & 1 / 1 & .54 / 1 & - & 1 / 1 & 1 / 1 & .36 / .63 & - & 1 & 1 & .92 & - \\
Circular & 1 & 1 & .76 & - & 1 / .47 & 1 / .26 & .88 / .61 & - & 1 / .94 & 1 / .52 & .83 / .83 & - & 1 / 1 & 1 / 1 & .77 / .83 & - & 1 / 1 & 1 / 1 & .72 / .5 & - & 1 / 1 & 1 / 1 & .72 / .77 & - & 1 & 1 & .76 & - \\ \hline
\end{tabular}}
\caption{Auto-GLOSH vs KNN and LOF across $k$ values in $[5, 10, 25, 50, 100]$: Precision@n obtained with global outliers (G), clumps (C), local outliers (L), and mixed outliers (M)}
\label{tab:autoGLOSH_vs_knn_lof}
\end{table*}

\begin{table}[t]
\centering
\resizebox{8.2cm}{1.8cm}{%
\begin{tabular}{|c|cccc|cccc|cccc|cccc|}
\hline
\multirow{2}{*}{\textbf{Datasets}} & \multicolumn{4}{c|}{\textbf{Auto-GLOSH}} & \multicolumn{4}{c|}{\textbf{ABOD}} & \multicolumn{4}{c|}{\textbf{COPOD}} & \multicolumn{4}{c|}{\textbf{*Best GLOSH}} \\ \cline{2-17} 
 & \textbf{G} & \textbf{C} & \textbf{L} & \textbf{M} & \textbf{G} & \textbf{C} & \textbf{L} & \textbf{M} & \textbf{G} & \textbf{C} & \textbf{L} & \textbf{M} & \textbf{G} & \textbf{C} & \textbf{L} & \textbf{M} \\ \hline
MVTec-AD\_zipper & 1 & 1 & 1 & 1 & 1 & 1 & 1 & 1 & 1 & 1 & 1 & 1 & 1 & 1 & 1 & 1 \\
HEPATITIS & 1 & 1 & 1 & 1 & 1 & 1 & 1 & 1 & 1 & 1 & 1 & 1 & 1 & 1 & 1 & 1 \\
LETTER & 1 & 1 & \textbf{.9} & 1 & 1 & 1 & .85 & 1 & 1 & 1 & .86 & 1 & 1 & 1 & .93 & 1 \\
PIMA & 1 & \textbf{1} & \textbf{.86} & 1 & 1 & .98 & .68 & 1 & 1 & \textbf{1} & .77 & 1 & 1 & 1 & .86 & 1 \\
STAMPS & \textbf{1} & \textbf{1} & .35 & 1 & .96 & .83 & \textbf{.5} & 1 & .96 & \textbf{1} & \textbf{.5} & 1 & 1 & 1 & .57 & 1 \\
VERTEBRAL & .9 & \textbf{1} & .7 & .95 & \textbf{.95} & .9 & \textbf{.78} & .95 & .9 & \textbf{1} & .64 & \textbf{1} & .9 & 1 & .7 & .95 \\
VOWELS & 1 & \textbf{1} & \textbf{.88} & \textbf{1} & 1 & \textbf{1} & \textbf{.88} & \textbf{1} & 1 & .99 & .53 & .97 & 1 & 1 & .92 & 1 \\
WDBC & 1 & \textbf{1} & \textbf{.66} & .96 & 1 & .91 & \textbf{.66} & .96 & 1 & \textbf{1} & .33 & \textbf{1} & 1 & 1 & 1 & .96 \\
WINE & 1 & 1 & \textbf{1} & 1 & 1 & 1 & \textbf{1} & 1 & 1 & 1 & .5 & 1 & 1 & 1 & 1 & 1 \\
WPBC & 1 & 1 & .8 & 1 & 1 & 1 & .8 & 1 & 1 & 1 & .8 & 1 & 1 & 1 & 1 & 1 \\
YEAST & 1 & .42 & .87 & .76 & 1 & .6 & .7 & .81 & 1 & \textbf{.79} & \textbf{.91} & \textbf{.92} & 1 & .46 & .89 & .83 \\
BREASTW & \textbf{1} & \textbf{1} & .22 & \textbf{1} & \textbf{1} & .63 & .36 & .89 & .97 & .97 & \textbf{.72} & \textbf{1} & 1 & 1 & .22 & 1 \\
CARDIO & 1 & .79 & .87 & .92 & 1 & .86 & .8 & .93 & 1 & \textbf{.9} & \textbf{.93} & \textbf{.97} & 1 & .85 & .89 & .94 \\
CARDIOTOCOGRAPHY & 1 & 1 & .89 & 1 & 1 & 1 & .86 & 1 & 1 & 1 & \textbf{.95} & 1 & 1 & 1 & .93 & 1 \\
20news\_3 & 1 & \textbf{1} & \textbf{.95} & \textbf{.98} & 1 & .98 & .91 & \textbf{.98} & 1 & .98 & .78 & .96 & 1 & 1 & .95 & 1 \\ \hline
Anisotropic & \textbf{1} & 1 & \textbf{.69} & - & 1 & .89 & .53 & - & \textbf{.73} & 1 & .3 & - & 1 & 1 & .76 & - \\
Banana & 1 & 1 & \textbf{.92} & - & 1 & 1 & .9 & - & 1 & 1 & .45 & - & 1 & 1 & .92 & - \\
Circular & 1 & 1 & \textbf{.76} & - & 1 & 1 & .61 & - & 1 & 1 & .5 & - & 1 & 1 & .76 & - \\ \hline
\textit{\textbf{* Winner count *}} & \multicolumn{4}{c|}{\textbf{22}} & \multicolumn{4}{c|}{9} & \multicolumn{4}{c|}{17} & \multicolumn{4}{c|}{-} \\ \hline
\end{tabular}}
\caption{Auto-GLOSH vs ABOD and COPOD: Precision@n (P@n) obtained with global outliers (G), clumps (C), local outliers (L), and mixed outliers (M). \emph{* Winner Count *} indicates the number of times a method achieves the highest P@n without tying with the other two methods.}
\label{tab:autoGLOSH_vs_abod_copod}
\end{table}
\raggedbottom

\noindent \textbf{Auto-GLOSH Vs. Parameter-free Methods.} In Table \ref{tab:autoGLOSH_vs_abod_copod},
Auto-GLOSH consistently achieves high P@n across most datasets compared to ABOD and COPOD. Similar to Table \ref{tab:autoGLOSH_vs_knn_lof}, there are instances, such as BREASTW with local outliers, where COPOD outperforms the best results of GLOSH. Consequently, it also outperforms Auto-GLOSH. 
As COPOD measures outlierness along each dimension, it is able to identify the outliers that deviate only along certain dimensions. However, we argue that our method is the most reliable compared to the state-of-the-art parameter-free methods, as it records the highest P@n for most datasets across all kinds of outliers. 

\noindent \textbf{Runtime Analysis.} Concerning the runtime of the algorithms presented in Table \ref{tab:runtime}, it is clear that by using \emph{CORE-SG} we can achieve results much faster than running HDBSCAN* at every $min_{pts}$ value up to $m_{max}$ (the naïve approach).
This faster running time is achieved as the base \emph{CORE-SG} graph is built just once, and the $MST$ for each $min_{pts}$ value is independently extracted from it.
Although COPOD displayed the best runtime, Auto-GLOSH obtained a similar performance, and both algorithms ran in less than a second. Although the runtime is comparable, Auto-GLOSH is still more reliable than COPOD in terms of outlier detection, as we reported in Table \ref{tab:autoGLOSH_vs_abod_copod}.
However, the runtime of ABOD is significantly higher than the other algorithms as its outlier score is computed w.r.t. all the points in the dataset.


\begin{table}[]
\centering
\resizebox{!}{0.9cm}{%
\begin{tabular}{|c|c|c|c|}
\hline
\textbf{Method} & \textbf{Global} & \textbf{Clumps} & \textbf{Local} \\ \hline
Auto-GLOSH & .0138 & .0141 & .0139 \\
\textit{+ CORE-SG} & .6127 & .6196 & .6041 \\ \hline
Naïve Auto-GLOSH & 4.448 & 4.619 & 4.508 \\ \hline
ABOD & 467.2 & 467.04 & 458.5 \\ \hline
COPOD & .0014 & .0015 & .0016 \\ \hline
\end{tabular}}
\caption{Comparing the average runtime (in seconds). The \emph{+ CORE-SG} adds the \emph{CORE-SG} construction time to that of Auto-GLOSH. The runtime is averaged over all the datasets for each outlier type.}
\label{tab:runtime}
\end{table}

\begin{table*}[t]\small
\centering
\resizebox{\textwidth}{2.0cm}{%
\begin{tabular}{|c|cccccccccccccccccccccccc|cccccccc|}
\hline
\multirow{4}{*}{\textbf{Dataset}} & \multicolumn{24}{c|}{\textbf{POLAR}} & \multicolumn{8}{c|}{\multirow{2}{*}{\textbf{*Best}}} \\ \cline{2-25}
 & \multicolumn{12}{c|}{\textbf{Knee}} & \multicolumn{12}{c|}{\textbf{Adjusted}} & \multicolumn{8}{c|}{} \\ \cline{2-33} 
 & \multicolumn{4}{c|}{\textbf{Recall}} & \multicolumn{4}{c|}{\textbf{F-measure}} & \multicolumn{4}{c|}{\textbf{G-Mean}} & \multicolumn{4}{c|}{\textbf{Recall}} & \multicolumn{4}{c|}{\textbf{F-measure}} & \multicolumn{4}{c|}{\textbf{G-Mean}} & \multicolumn{4}{c|}{\textbf{F-measure}} & \multicolumn{4}{c|}{\textbf{G-Mean}} \\ \cline{2-33} 
 & \textbf{G} & \textbf{C} & \textbf{L} & \multicolumn{1}{c|}{\textbf{M}} & \textbf{G} & \textbf{C} & \textbf{L} & \multicolumn{1}{c|}{\textbf{M}} & \textbf{G} & \textbf{C} & \textbf{L} & \multicolumn{1}{c|}{\textbf{M}} & \textbf{G} & \textbf{C} & \textbf{L} & \multicolumn{1}{c|}{\textbf{M}} & \textbf{G} & \textbf{C} & \textbf{L} & \multicolumn{1}{c|}{\textbf{M}} & \textbf{G} & \textbf{C} & \textbf{L} & \textbf{M} & \textbf{G} & \textbf{C} & \textbf{L} & \multicolumn{1}{c|}{\textbf{M}} & \textbf{G} & \textbf{C} & \textbf{L} & \textbf{M} \\ \hline
MVTec-AD\_zipper & 1 & 1 & 1 & \multicolumn{1}{c|}{1} & .60 & .49 & .27 & \multicolumn{1}{c|}{.66} & .96 & .94 & .97 & \multicolumn{1}{c|}{.95} & 1 & 1 & 1 & \multicolumn{1}{c|}{1} & 1 & .87 & .66 & \multicolumn{1}{c|}{1} & 1 & .99 & .99 & 1 & 1 & 1 & 1 & \multicolumn{1}{c|}{1} & 1 & 1 & 1 & 1 \\
HEPATITIS & 1 & 1 & 1 & \multicolumn{1}{c|}{1} & .66 & .75 & .11 & \multicolumn{1}{c|}{.85} & .97 & .98 & .57 & \multicolumn{1}{c|}{.98} & 1 & 1 & 0 & \multicolumn{1}{c|}{1} & 1 & 1 & 0 & \multicolumn{1}{c|}{.92} & 1 & 1 & 0 & .99 & 1 & 1 & 1 & \multicolumn{1}{c|}{1} & 1 & 1 & 1 & 1 \\
LETTER & 1 & 1 & 1 & \multicolumn{1}{c|}{1} & .61 & .61 & .65 & \multicolumn{1}{c|}{.78} & .96 & .96 & .97 & \multicolumn{1}{c|}{.97} & 1 & 1 & .85 & \multicolumn{1}{c|}{1} & .98 & .94 & .91 & \multicolumn{1}{c|}{1} & .99 & .99 & .92 & 1 & 1 & 1 & .92 & \multicolumn{1}{c|}{1} & 1 & 1 & .98 & 1 \\
PIMA & 1 & 1 & 1 & \multicolumn{1}{c|}{1} & .7 & .61 & .29 & \multicolumn{1}{c|}{.65} & .97 & .96 & .88 & \multicolumn{1}{c|}{.94} & 1 & 1 & .9 & \multicolumn{1}{c|}{1} & .94 & .96 & .88 & \multicolumn{1}{c|}{.67} & .99 & .99 & .95 & .94 & 1 & 1 & .93 & \multicolumn{1}{c|}{1} & 1 & 1 & .99 & 1 \\
STAMPS & 1 & 1 & 1 & \multicolumn{1}{c|}{1} & .34 & .41 & .40 & \multicolumn{1}{c|}{.61} & .9 & .92 & .93 & \multicolumn{1}{c|}{.92} & 1 & 1 & .35 & \multicolumn{1}{c|}{1} & .73 & .79 & .41 & \multicolumn{1}{c|}{.66} & .98 & .98 & .59 & .94 & 1 & 1 & .53 & \multicolumn{1}{c|}{1} & 1 & 1 & .95 & 1 \\
VERTEBRAL & 1 & 1 & 1 & \multicolumn{1}{c|}{1} & .34 & .37 & .35 & \multicolumn{1}{c|}{.55} & .9 & .91 & .91 & \multicolumn{1}{c|}{.91} & 1 & 1 & .8 & \multicolumn{1}{c|}{1} & .68 & .71 & .61 & \multicolumn{1}{c|}{.85} & .97 & .98 & .87 & .98 & .95 & 1 & .77 & \multicolumn{1}{c|}{.97} & .99 & 1 & .96 & .99 \\
VOWELS & 1 & 1 & 1 & \multicolumn{1}{c|}{1} & .73 & .46 & .35 & \multicolumn{1}{c|}{.66} & .98 & .94 & .96 & \multicolumn{1}{c|}{.95} & 1 & 1 & .96 & \multicolumn{1}{c|}{1} & .95 & .87 & .74 & \multicolumn{1}{c|}{1} & .99 & .99 & .97 & 1 & 1 & 1 & .92 & \multicolumn{1}{c|}{1} & 1 & 1 & .99 & 1 \\
WDBC & 1 & 1 & 1 & \multicolumn{1}{c|}{1} & .36 & .33 & .08 & \multicolumn{1}{c|}{.46} & .9 & .89 & .91 & \multicolumn{1}{c|}{.90} & 1 & 1 & 1 & \multicolumn{1}{c|}{1} & .85 & .83 & .3 & \multicolumn{1}{c|}{.68} & .99 & .99 & .98 & .96 & 1 & 1 & .85 & \multicolumn{1}{c|}{.98} & 1 & 1 & .99 & .99 \\
WINE & 1 & 1 & 1 & \multicolumn{1}{c|}{1} & .44 & .41 & .26 & \multicolumn{1}{c|}{.51} & .93 & .92 & .95 & \multicolumn{1}{c|}{.91} & 1 & 1 & 1 & \multicolumn{1}{c|}{1} & 1 & .92 & .5 & \multicolumn{1}{c|}{1} & 1 & .99 & .98 & 1 & 1 & 1 & 1 & \multicolumn{1}{c|}{1} & 1 & 1 & 1 & 1 \\
WPBC & 1 & 1 & 1 & \multicolumn{1}{c|}{1} & .18 & .34 & .25 & \multicolumn{1}{c|}{.45} & .9 & .89 & .89 & \multicolumn{1}{c|}{.89} & 1 & 1 & 1 & \multicolumn{1}{c|}{1} & .54 & .89 & .66 & \multicolumn{1}{c|}{.83} & .98 & .99 & .98 & .98 & 1 & 1 & .88 & \multicolumn{1}{c|}{1} & 1 & 1 & .99 & 1 \\
YEAST & 1 & 1 & .97 & \multicolumn{1}{c|}{1} & .32 & .31 & .33 & \multicolumn{1}{c|}{.43} & .89 & .88 & .88 & \multicolumn{1}{c|}{.85} & .75 & .4 & .97 & \multicolumn{1}{c|}{.83} & .69 & .4 & .77 & \multicolumn{1}{c|}{.77} & .86 & .62 & .97 & .89 & .73 & .56 & .91 & \multicolumn{1}{c|}{.77} & .96 & .95 & .98 & .95 \\
BREASTW & 1 & 1 & 1 & \multicolumn{1}{c|}{1} & .23 & .21 & .16 & \multicolumn{1}{c|}{.93} & .82 & .8 & .71 & \multicolumn{1}{c|}{.99} & 1 & 1 & .72 & \multicolumn{1}{c|}{1} & .51 & .46 & .31 & \multicolumn{1}{c|}{.65} & .95 & .94 & .78 & .94 & 1 & 1 & .36 & \multicolumn{1}{c|}{1} & 1 & 1 & .83 & 1 \\
CARDIO & 1 & 1 & 1 & \multicolumn{1}{c|}{1} & .42 & .37 & .39 & \multicolumn{1}{c|}{.56} & .93 & .91 & .92 & \multicolumn{1}{c|}{.91} & 1 & .71 & .98 & \multicolumn{1}{c|}{.91} & .94 & .76 & .73 & \multicolumn{1}{c|}{.92} & .99 & .83 & .97 & .95 & 1 & .85 & .88 & \multicolumn{1}{c|}{.94} & 1 & .99 & .98 & .99 \\
CARDIOTOCOGRAPHY & 1 & 1 & 1 & \multicolumn{1}{c|}{1} & .65 & .55 & .48 & \multicolumn{1}{c|}{.56} & .97 & .95 & .94 & \multicolumn{1}{c|}{.91} & 1 & 1 & .79 & \multicolumn{1}{c|}{1} & .96 & .96 & .84 & \multicolumn{1}{c|}{.98} & .99 & .99 & .98 & .99 & 1 & 1 & .91 & \multicolumn{1}{c|}{1} & 1 & 1 & .98 & 1 \\
20news\_3 & 1 & 1 & 1 & \multicolumn{1}{c|}{1} & .45 & .43 & .43 & \multicolumn{1}{c|}{.61} & .93 & .93 & .94 & \multicolumn{1}{c|}{.93} & 1 & 1 & 1 & \multicolumn{1}{c|}{1} & .79 & .78 & .75 & \multicolumn{1}{c|}{.91} & .98 & .98 & .98 & .99 & 1 & 1 & .97 & \multicolumn{1}{c|}{.99} & 1 & 1 & .99 & .99 \\ \hline
Anisotropic & 1 & 1 & 1 & \multicolumn{1}{c|}{-} & .3 & .21 & .24 & \multicolumn{1}{c|}{-} & .93 & .88 & .93 & \multicolumn{1}{c|}{-} & 1 & 1 & .92 & \multicolumn{1}{c|}{-} & .95 & .86 & .6 & \multicolumn{1}{c|}{-} & .99 & .99 & .94 & - & 1 & 1 & .74 & \multicolumn{1}{c|}{-} & 1 & 1 & .98 & - \\
Banana & 1 & 1 & .92 & \multicolumn{1}{c|}{-} & .33 & .45 & .55 & \multicolumn{1}{c|}{-} & .93 & .96 & .94 & \multicolumn{1}{c|}{-} & 1 & 1 & .53 & \multicolumn{1}{c|}{-} & .92 & 1 & .7 & \multicolumn{1}{c|}{-} & .99 & 1 & .73 & - & 1 & 1 & .96 & \multicolumn{1}{c|}{-} & 1 & 1 & .96 & - \\
Circular & 1 & 1 & .76 & \multicolumn{1}{c|}{-} & .77 & .76 & .68 & \multicolumn{1}{c|}{-} & .99 & .99 & .87 & \multicolumn{1}{c|}{-} & .89 & 1 & .3 & \multicolumn{1}{c|}{-} & .94 & 1 & .47 & \multicolumn{1}{c|}{-} & .94 & 1 & .55 & - & 1 & 1 & .86 & \multicolumn{1}{c|}{-} & 1 & 1 & .87 & - \\ \hline
\end{tabular}}
\caption{Evaluating POLAR on datasets with global outliers (G), clumps (C), local outliers (L), and mixed outliers (M). We compare with the best performance that can be achieved across all possible thresholds.}
\label{tab:polar}
\end{table*}
\subsection{POLAR}
\label{sub_sec:result_polar}

\noindent \textbf{Setup.} 
Here, we evaluate POLAR by answering two questions:
(I) does the adjusted threshold yield better results than the ``knee'' GLOSH score? and (II) does the estimated thresholds match the performance with the best threshold? We evaluate POLAR using recall, F-measure, and
G-Mean \cite{aurelio2019learning}.

\noindent \textbf{Results.} In Table \ref{tab:polar}, w.r.t. F-measure and G-Mean, overall, the Adjusted Threshold yields a performance that is close to the best in most cases. However, for most cases, taking the knee GLOSH score as the threshold gives a perfect recall, whereas the adjusted threshold shows comparatively lower recall (higher false negatives) in many cases (prominent for local outliers). However, the adjusted threshold mostly gives a higher F-measure as it increases the precision (reduces false positives). Therefore, when finding true positives is more important than avoiding false positives, the unadjusted knee threshold may be preferred.

\label{tex:experimental_analysis}

\section{Conclusion}
\label{sec:conc;usion}

This paper successfully addresses two critical challenges in GLOSH: finding the ``best'' $min_{pts}$ value and a suitable threshold for labeling inliers and ``potential outliers''. Both the proposed approaches, Auto-GLOSH for $min_{pts}$ selection and POLAR for automatic labeling, closely approximate the best or nearly best performance across various datasets and achieve parameter-free unsupervised outlier detection.

\label{tex:conclusion}

\bibliographystyle{IEEEtran}
\bibliography{references.bib}

\end{document}